\def\year{2022}\relax
\documentclass[letterpaper]{article} % DO NOT CHANGE THIS
\usepackage{aaai22}  % DO NOT CHANGE THIS
\usepackage{times}  % DO NOT CHANGE THIS
\usepackage{helvet}  % DO NOT CHANGE THIS
\usepackage{courier}  % DO NOT CHANGE THIS
\usepackage[hyphens]{url}  % DO NOT CHANGE THIS
\usepackage{graphicx} % DO NOT CHANGE THIS
\usepackage{natbib}  % DO NOT CHANGE THIS AND DO NOT ADD ANY OPTIONS TO IT
\usepackage{caption} % DO NOT CHANGE THIS AND DO NOT ADD ANY OPTIONS TO IT
\DeclareCaptionStyle{ruled}{labelfont=normalfont,labelsep=colon,strut=off} % DO NOT CHANGE THIS
\usepackage{algorithm}
\usepackage[noend]{algpseudocode}

\usepackage{hyperref}       % hyperlinks
\usepackage{booktabs}       % professional-quality tables
\usepackage{amsfonts}       % blackboard math symbols
\usepackage{nicefrac}       % compact symbols for 1/2, etc.
\usepackage{microtype}      % microtypography
\usepackage{amsmath}
\usepackage{bm}
\usepackage{subcaption}
\usepackage{multirow}
\usepackage{enumitem}
\usepackage[toc]{appendix}

\usepackage{listings}
\usepackage{xcolor}

\usepackage{tikz}
\usetikzlibrary{decorations.pathreplacing,calc}
\newcommand{\tikzmark}[1]{\tikz[overlay,remember picture] \node (#1) {};}

\newcommand*{\AddNote}[4]{%
    \begin{tikzpicture}[overlay, remember picture]
        \draw [decoration={brace,amplitude=0.5em,mirror},decorate, thick]
            ($(#3)!(#1.north)!($(#3)-(0,1)$)$) --  
            ($(#3)!(#2.south)!($(#3)-(0,1)$)$)
                node [align=center, text width=2.5cm, pos=0.5, anchor=east] {#4};
    \end{tikzpicture}
}%

\definecolor{codegreen}{rgb}{0,0.6,0}
\definecolor{codegray}{rgb}{0.5,0.5,0.5}
\definecolor{codepurple}{rgb}{0.58,0,0.82}
\definecolor{backcolour}{rgb}{0.95,0.95,0.92}

\lstdefinestyle{mystyle}{
    backgroundcolor=\color{backcolour},   
    commentstyle=\color{codegreen},
    keywordstyle=\color{magenta},
    numberstyle=\tiny\color{codegray},
    stringstyle=\color{codepurple},
    basicstyle=\ttfamily\footnotesize,
    breakatwhitespace=false,         
    breaklines=true,                 
    captionpos=b,                    
    keepspaces=true,                 
    numbers=left,                    
    numbersep=5pt,                  
    showspaces=false,                
    showstringspaces=false,
    showtabs=false,                  
    tabsize=2
}

\DeclareMathOperator{\argmin}{\arg\!\min}
\newcommand\defeq{\mathrel{\overset{\makebox[0pt]{\mbox{\normalfont\tiny\sffamily def}}}{=}}}

\usepackage{newfloat}
\usepackage{listings}
\floatstyle{ruled}
\newfloat{listing}{tb}{lst}{}
\floatname{listing}{Listing}
\title{PUMA: Performance Unchanged Model Augmentation for Training Data Removal}

\author {
    Ga Wu,
    Masoud Hashemi,
    Christopher Srinivasa
}
\affiliations {
    Borealis AI\\
     \{ga.wu, masoud.hashemi, christopher.srinivasa\}@borealisai.com
}

\usepackage{bibentry}
\usepackage{color}
\usepackage{xcolor}

\begin{document}

\maketitle

\begin{abstract}
Preserving the performance of a trained model while removing unique characteristics of marked training data points is challenging. Recent research usually suggests retraining a model from scratch with remaining training data or refining the model by reverting the model optimization on the marked data points. Unfortunately, aside from their computational inefficiency, those approaches inevitably hurt the resulting model's generalization ability since they remove not only unique characteristics but also discard shared (and possibly contributive) information. To address the performance degradation problem, this paper presents a novel approach called Performance Unchanged Model Augmentation~(PUMA). The proposed PUMA framework explicitly models the influence of each training data point on the model's generalization ability with respect to various performance criteria. It then complements the negative impact of removing marked data by reweighting the remaining data optimally. To demonstrate the effectiveness of the PUMA framework, we compared it with multiple state-of-the-art data removal techniques in the experiments, where we show the PUMA can effectively and efficiently remove the unique characteristics of marked training data without retraining the model that can 
1) fool a membership attack,
and 2) resist performance degradation.
In addition, as PUMA estimates the data importance during its operation, we show it could serve to  debug mislabelled data points more efficiently than existing approaches.
\end{abstract}

\section{Introduction}
%Preserving performance for the machine learning model when removing sensitive training data points is a critical and challenging. 
As many countries and territories become increasingly concerned with personal data protection, the corresponding protection regulations\footnote{CCPA in California, GDPR in Europe, PIPEDA in Canada, LGPD in Brazil, and NDBS in Australia.} entitle individuals to revoke their authorization of using their data for data analysis and machine learning~(ML) model training. While retraining ML models by removing marked data points is a feasible solution, frequent data removal requests inevitably put enormous computational pressure on the infrastructures responsible for real-time ML services. Furthermore, cumulative data loss results in quick performance degradation. Hence, effectively eliminating data's unique characteristics while preserving model performance is a critical and challenging research question.

In the literature, a few initial works attempted to address the data removal challenge.
For example, \cite{GinartGVZ19} devised a general notion of \textit{removal efficiency} and proposed two model-specific data removal algorithms (for k-means clustering models). Similarly, \cite{GuoGHM20} introduced a notion of \textit{Certified Removal} and verified the effectiveness of their data removal approach on linear classifiers. However, those methods usually focus on specific ML algorithms and are hard to generalize to deep neural networks that dominate the latest ML research and applications.
\cite{Bourtoule2019}, alternatively, proposed a data removal-friendly model by ensembling multiple ML models trained on disjoint data partitions. As such, the data removal operation would only involve a sub-model. \cite{Vijay2020} proposed a more generalized single-model solution by explicitly estimating the contribution (gradients) of each training data point as an additive function. Unfortunately, such approaches require high costs; maintaining many sub-models and tracking the model training process are barely feasible for real-world applications. In addition, existing data removal works merely pay attention to the performance degradation problem when removing marked data points. While \cite{GinartGVZ19}'s criterion includes a constraint such as performance of the resulting model should not be worse than that of a model trained from scratch with remaining data, it does not intend to preserve the performance of the original model.

In this paper, we propose a novel approach, Performance Unchanged Model Augmentation~(PUMA), to efficiently erase the unique characteristics of marked data points from a trained model without causing performance degradation. In particular, the proposed PUMA framework explicitly models the influence of each training data point on the model with respect to various performance criteria (that are not necessarily the model training objectives). It then complements the negative impact of removing marked data by reweighting the remaining data points sparsely and optimally through a constrained optimization. Consequently, PUMA can preserve model performance by linearly patching the original model via reweighting operation while eliminating unique characteristics of marked data points. In the experiments, we compare PUMA with existing data removal approaches and show that PUMA has two desired properties: 1) It can successfully fool a membership attack~\cite{shokri2017membership}, 2) It can resist performance degradation.
\section{Preliminary and Related Works}
Before proceeding, we review existing related data removal approaches which inspired this work. We also briefly describe the influence function to facilitate our description in the main content. Finally, we list several information leaking attack approaches that can be used to test the effectiveness of data removal in the existing literature.
\subsection{Data Removal Approaches}
Removing training data from models has a long research history that can be tracked back to the era of support vector machines. \cite{CauwenberghsP00} proposed a decremented unlearning approach, called Leave-One-Out~(LOO), to gradually remove marked training data points from trained SVM model. By examining the margin of the data points, LOO could significantly reduce the computational effort of data removal. Later, \cite{KarasuyamaT09} extended the decremental unlearning approach to support simultaneous addition and/or removal of multiple data points through multi-parametric programming. Following the same line of research, \cite{TsaiLL14} proposed a warm-up based unlearning approach that is effective on multiple linear machine learning models. Lastly, \cite{GinartGVZ19} payed attention to unsupervised learning tasks where it presented two model-specific data removal algorithms for k-means clustering models.

Recent research~\cite{Vijay2020} stated that the previously mentioned approaches are not suitable to work on deep network models where the contribution of individual training data points are intractable to compute exactly and analytically. To mitigate the computational cost of retraining a new model from scratch, \cite{Bourtoule2019} suggested training multiple models on disjoint data partitions so that retraining is limited to small groups of sub-models. Alternatively, \cite{Vijay2020} presented \textit{Amnesiac training} which tracks contribution of each training batch (a set of data points) during the model training. When a batch is marked as to be removed, the operation is simply a subtraction between model parameters and data contribution.

While the existing approaches show remarkable achievement on improving efficiency of removing data points from a trained model, we note that they underestimated two critical criteria of data removal tasks: 1) The data removal approach should maintain model stability and protect against performance degradation. 2) The data removal approach should minimize the overall computational cost instead of only looking at the cost of the data removal operation. More specifically, training multiple models or tracking gradients of every training epoch is undesired in practice. All of the above observations motivated our work on proposing Performance Unchanged Model Augmentation~(PUMA) in this paper. 

\subsection{Influence Function for Prediction Explanation}
An influence function is a limit equation which estimates the prediction changes of a model when its inputs are perturbed. In statistics, the influence function is similar to the G\^{a}teaux derivative, but it can exist even when the G\^{a}teaux derivative does not exist for a particular model.

Recently, the influence function was used to explain the prediction of complex machine learning models as it can reveal the impact of training data point $(\mathbf{x}_k, y_k)$ on the test example $(\mathbf{x}_j, y_j)$'s predictions~\cite{KohL17} such that
\begin{equation}
\begin{aligned}
    &\mathcal{I}_{\textit{up},\textit{loss}}\left((\mathbf{x}_k, y_k), (\mathbf{x}_j, y_j)\right) \stackrel{\text{def}}{=} \frac{d \mathcal{L}(\mathbf{x}_j,y_j,\theta)}{d\epsilon}\Bigr|_{\epsilon=0}\\
    &=\!-\nabla_\theta \mathcal{L}(\mathbf{x}_j,\!y_j,\!\theta) \!\left(\frac{1}{m}\!\sum_{i=1}^m\!\nabla_{\theta}^{2}\mathcal{L}(\mathbf{x}_i,\!y_i,\!\theta) \right)^{\!-1}\!\!\!\!\!\nabla_\theta \mathcal{L}(\mathbf{x}_k,\!y_k,\!\theta),
\end{aligned}
\label{eq:influence_explanation}
\end{equation}
where $\mathcal{L}$ denotes the loss function for the individual data point, and $\epsilon$ denotes the degree of perturbation on the data $k$. By computing Equation~\ref{eq:influence_explanation} for all training data points $k$, we can summarize a training data importance rank for a particular test sample $j$. 

Naturally, if we can explain the model prediction based on its training data points, we can also refine the model prediction by perturbing those data points. Based on this idea, \cite{GuoGHM20} proposed a data removal approach that leverages the Newton method and influence function. However, their solution is defined for a linear model, making it hard to verify its performance on complex models. 

In this work, we will also leverage the influence function. The critical difference between our work and \cite{GuoGHM20} is two-folds: First, our objective is to let the modified model preserve the original model's performance after data removal rather than passively monitoring whether the modified model can produce near identical predictions against a model trained on the remaining data from scratch. When a huge number of data points are requested to remove, the difference between these two objectives is significant; training new model from scratch with insufficient data points may not reach a desirable performance. Second, the proposed approach modifies all trainable parameters of the model while \cite{GuoGHM20} only adjusts the linear decision making layer which does not eliminate unique characteristics of the removed data points (since the representations are learned with the knowledge of the removed data points). 
% (and retraining the model from scratch if the performance is not preserved)

\subsection{Data Privacy Protection and Membership Attacks}
In terms of evaluating the effectiveness of data removal approaches, previous research \cite{Vijay2020} suggested leveraging information leaking attacks~\cite{homer2008resolving,DworkSSUV15,FredriksonJR15,YeomGFJ18} to check if the data characteristics are indeed removed from a trained model. Specifically, it is suggested that the \textit{membership attack}~\cite{homer2008resolving} could reveal whether a particular data point is present in training a model, which is an ideal reference to see the difference of attacks before and after the data removal operation. In the literature, there are various membership attack algorithms~\cite{shokri2017membership,Milad18,YeomGFJ18} since the concept was introduced by~\cite{homer2008resolving}.

In this paper, we will follow the track of previous works and conduct membership attack experiments to show the effectiveness of our model in the experiments.
\section{Performance Unchanged Model Augmentation}
% Problem Setting.
% 1. target model
% 2. sensitive data points
% 3. general training data points.
% 4. we want to remove characteristic of sensitive data points but reduce its influence on model performance.
Given a machine learning model $f_{\theta_{\textit{org}}}$ learned on training data set $D_{\textit{tn}}$, we aim to remove the unique characteristics of marked data points $D_{\textit{mk}} \subset D_{\textit{tn}}$ from the model by updating model parameters $\theta_{\textit{org}} \rightarrow \theta_{\textit{mod}}$ without seriously hurting its prediction performance with respect to various performance criteria $\mathcal{C}$ (or $\mathcal{L}_\textit{c}$ for an individual sample) such that
\begin{equation}
    \Bigr\lvert \underbrace{\frac{1}{|D_{\textit{tn}}|}\!\sum_{i=1}^{|D_{\textit{tn}}|}\! \mathcal{L}_{\textit{c}}(\mathbf{x}_i, y_i, \theta_{\textit{mod}})}_{\mathcal{C}(\theta_{\textit{mod}})}\!-\!\underbrace{\frac{1}{|D_{\textit{tn}}|}\!\sum_{i=1}^{|D_{\textit{tn}}|}\!\mathcal{L}_{\textit{c}}(\mathbf{x}_i, y_i, \theta_{\textit{org}})}_{\mathcal{C}(\theta_{\textit{org}})} \Bigr\rvert\!\leq\!\delta,
\label{eq:performance_gap}
\end{equation}
where $\delta$ is a small change in performance. In particular, we are interested in preserving overall performance rather than being concerned with a shift in an individual prediction.

\subsection{Influence of Training Data}
To tackle the data removal task defined above, we first need to reveal the underlining causal relation between training data perturbation and model performance variation. Specifically, in this section, we clarify two aspects of this connection: 1) How the training data changes would impact model parameters, and 2) How the parameter changes would impact the model performance with respect to specific criteria $\mathcal{C}$.

\subsubsection{Parameter as Linear Function of Data Contributions}
We start by analyzing how perturbing the training dataset would impact the model parameter changes via the influence function. 

Let us assume the model parameter $\theta_{\textit{org}}$ is the optimal solution of the (original) training objective $\mathcal{J}_{\textit{org}}$
\begin{equation}
\begin{aligned}
    &\theta_{\textit{org}} = \underset{\theta}{\argmin} ~\mathcal{J}_{\textit{org}} (\theta)=\underset{\theta}{\argmin}\frac{1}{|D_{\textit{tn}}|}\!\sum_{i=1}^{|D_{\textit{tn}}|}\!\mathcal{L}_{\textit{t}}(\mathbf{x}_i, y_i, \theta)
\end{aligned}
\label{eq:optimal_original}
\end{equation}
and $\theta_{\textit{mod}}$ is the optimal solution of a modified objective $\mathcal{J}_{\textit{mod}}$
\begin{equation}
\begin{aligned}
    &\theta_{\textit{mod}} = \underset{\theta}{\argmin} ~\mathcal{J}_{\textit{mod}} (\theta)=\\
    &\!\underset{\theta}{\argmin}\underbrace{\frac{1}{|D_{\textit{tn}}|}\!\sum_{i=1}^{|D_{\textit{tn}}|}\!\mathcal{L}_{\textit{t}}(\mathbf{x}_i, y_i, \theta)}_{\mathcal{J}_{\textit{org}} (\theta)}\!+\!\underbrace{\frac{1}{|D_{\textit{up}}|}\!\sum_{j=1}^{|D_{\textit{up}}|}\! \lambda_j\mathcal{L}_{\textit{t}}(\mathbf{x}_j, y_j, \theta)}_{\mathcal{J}_{\textit{add}} (\theta)}
\end{aligned}
\label{eq:optimal_modification}
\end{equation}
that optimizes an additional weighted objective $\mathcal{J}_{\textit{add}}$ on a subset of training data points $D_{\textit{up}} \subseteq D_{\textit{tn}}$, where $\mathcal{L}_{\textit{t}}$ denotes individual prediction loss\footnote{Training loss $\mathcal{L}_{\textit{t}}$ is not necessarily identical to the performance criterion loss  $\mathcal{L}_{\textit{c}}$ defined in Equation~\ref{eq:performance_gap}.} and $\bm{\lambda}\in {\rm I\!R}^{|D_{\textit{up}}|}$ denotes the weight vector of upweighted data points. 

When the values of weights $\bm{\lambda}$ are negligibly small, the derivative of the modified objective $\mathcal{J}_{\textit{mod}}$ with respect to its optimal parameters $\theta_{\textit{mod}}$ could be Taylor expanded at the local anchor $\theta_{\textit{org}}$ such that
\begin{equation}
\begin{aligned}
    & \underbrace{\nabla \mathcal{J}_{mod}(\theta_{\textit{mod}})}_{\approx 0}\!\approx\!\nabla \mathcal{J}_{mod}(\theta_{\textit{org}}) + \nabla^2 \mathcal{J}_{mod}(\theta_{\textit{org}})(\theta_{\textit{mod}}\!-\!\theta_{\textit{org}})\\
    &\approx\!\underbrace{\nabla \mathcal{J}_{\textit{org}}(\theta_{\textit{org}})}_{\approx 0} + \nabla \mathcal{J}_{add}(\theta_{\textit{org}}) + \nabla^2 \mathcal{J}_{mod}(\theta_{\textit{org}}) (\theta_{\textit{mod}}\!-\!\theta_{\textit{org}}).\\
\end{aligned}
\label{eq:optimal_parameter_gap}
\end{equation}
Since the both $\theta_{\textit{mod}}$ and $\theta_{\textit{org}}$ are optimal solutions with respect to their corresponding objective functions $\nabla \mathcal{J}_{mod}(\theta)$ and $\nabla \mathcal{J}_{org}(\theta)$ (whose derivatives are $0$s), the Equation~\ref{eq:optimal_parameter_gap} yields a difference between the two optimal solution $\theta_{\textit{mod}}$ and $\theta_{\textit{org}}$ such that
\begin{equation}
    \theta_{\textit{mod}} - \theta_{\textit{org}} \defeq -\left(\nabla^2 \mathcal{J}_{\textit{org}}(\theta_{\textit{org}})\right)^{-1}\nabla \mathcal{J}_{\textit{add}}(\theta_{\textit{org}}),
\label{eq:parameter_difference_original}
\end{equation}
where we relaxed the Hessian matrix $\nabla^2 \mathcal{J}_{mod}(\theta_{\textit{org}})$ to $\nabla^2 \mathcal{J}_{org}(\theta_{\textit{org}})$. There are multiple justifications for such relaxation. First, since the $\bm{\lambda}$s are set to be small values, such a setting makes the difference of these second order derivatives insignificant. Second, in practice, computing the Hessian matrix (or Hessian Vector Product described later) is usually an iterative and stochastic process which introduces larger noise than the relaxation we introduced here. It is worth to mention that the expression in  Equation~\ref{eq:parameter_difference_original} aligns with previous influence function work~\cite{KohL17} when $\bm{\lambda}$ is restricted as a one-hot vector (that only upweights a single data point). In our implementation, we compute HVP approximation in the same way as described in ~\cite{KohL17}.

By expanding the derivative of the additive perturbation term $\nabla \mathcal{J}_{add}(\theta_{\textit{org}})$, we can convert the Equation~\ref{eq:parameter_difference_original} to a linear function of the perturbation weight $\bm{\lambda}$ as follows:
\begin{equation}
    \theta_{\textit{mod}}\!-\!\theta_{\textit{org}} = -\!\sum_{j=1}^{|D_{\textit{up}}|}\!\lambda_j\!\left(\nabla^2 \mathcal{J}_{\textit{org}}(\theta_{\textit{org}})\right)^{\!-\!1}\nabla \mathcal{L}_{\textit{t}}(\mathbf{x}_j,\!y_j,\!\theta_{\textit{org}}).
\label{eq:local_parameter_difference}
\end{equation}
Indeed, with trained model whose parameter $\theta_{\textit{org}}$ is fixed, both the Hessian matrix $\nabla^2 \mathcal{J}_{\textit{org}}(\theta_{\textit{org}})$ and gradient vector $\nabla \mathcal{L}(\mathbf{x}_j, y_j, \theta_{\textit{org}})$ are constant for the fixed set of upweighted data points $D_{\textit{up}}$.

\begin{figure*}[t]
     \centering
     \begin{subfigure}[b]{0.33\linewidth}
         \centering
         \includegraphics[width=\linewidth]{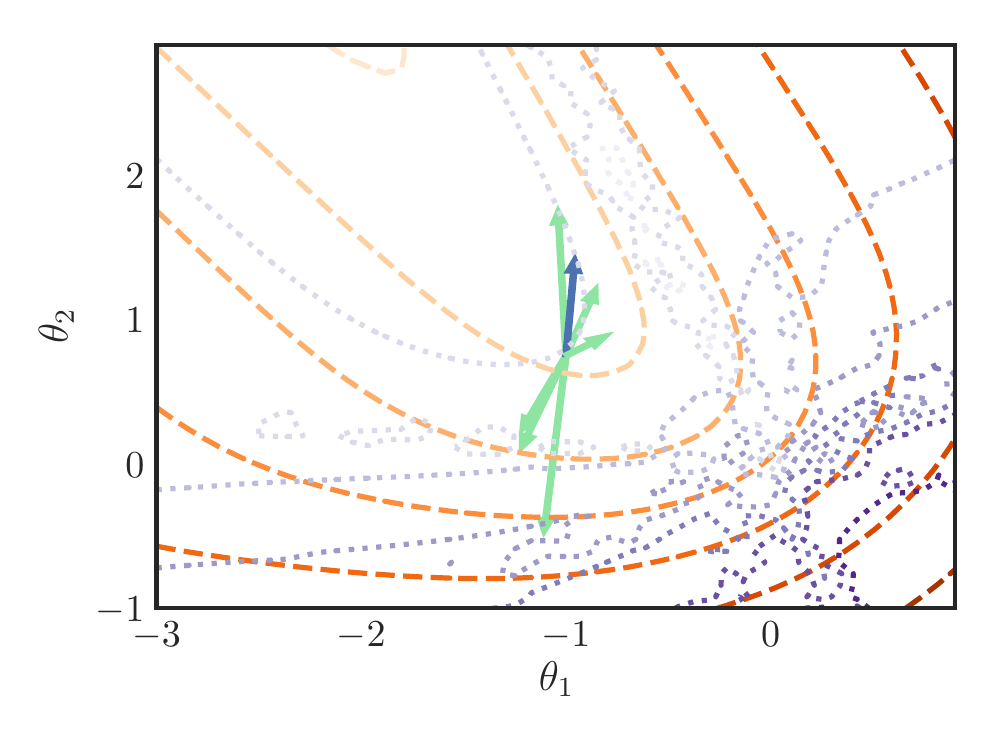}
         \caption{Before Removal}
     \end{subfigure}
     \begin{subfigure}[b]{0.33\linewidth}
         \centering
         \includegraphics[width=\linewidth]{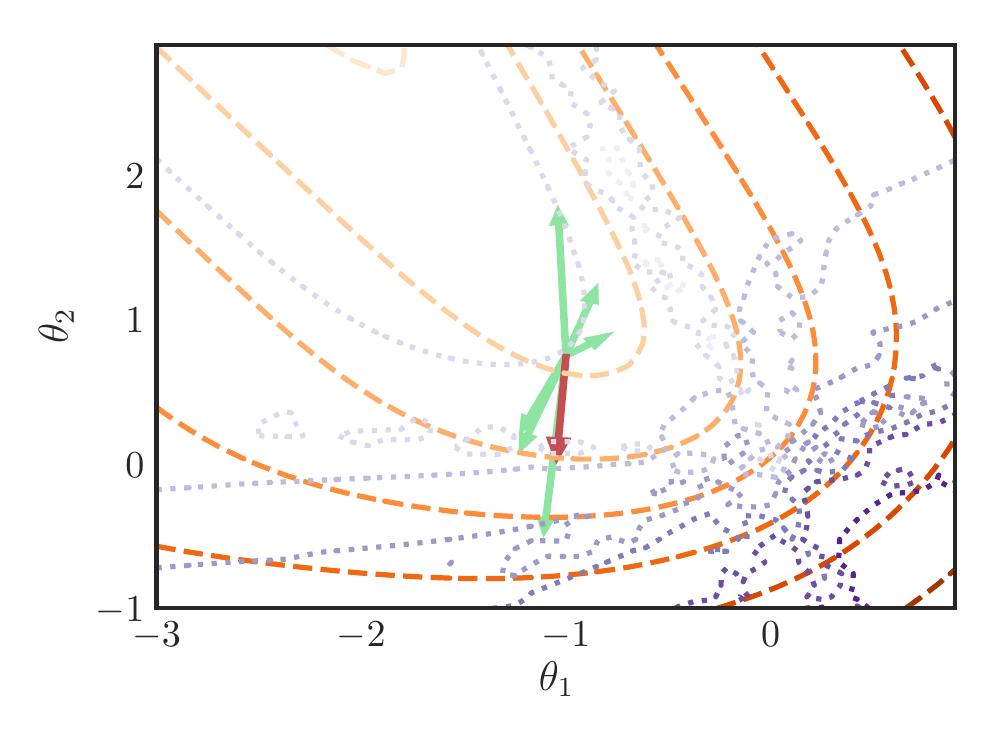}
         \caption{Naive Influence Removal}
     \end{subfigure}
     \begin{subfigure}[b]{0.33\linewidth}
        \centering
        \includegraphics[width=\linewidth]{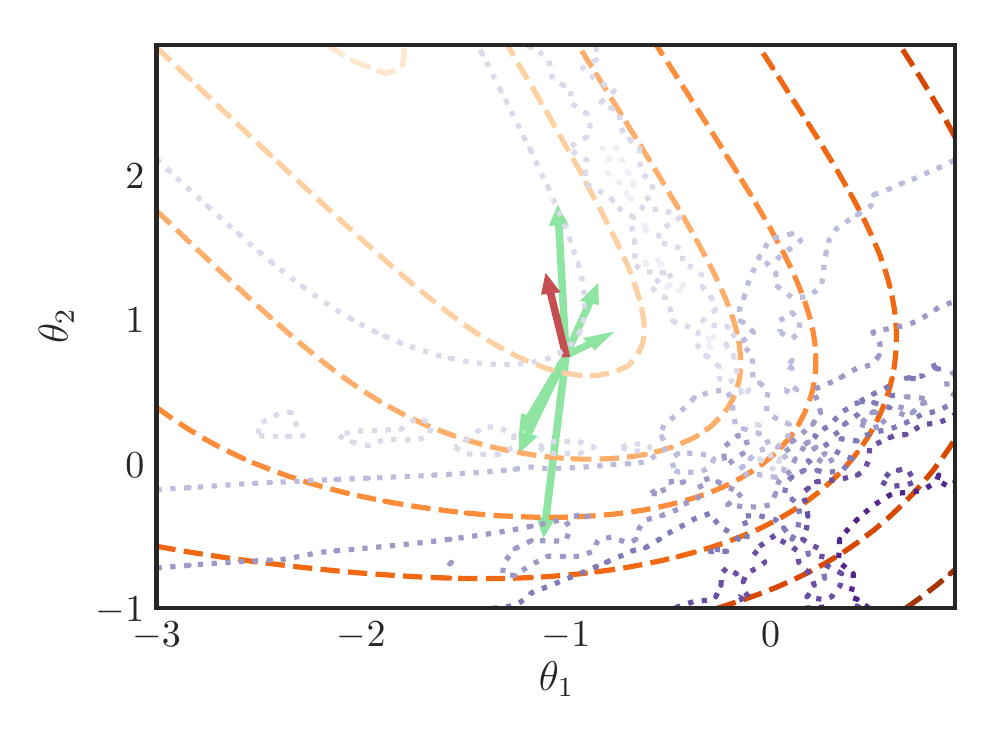}
         \caption{PUMA}
     \end{subfigure}
    \caption{\textbf{Projection Direction Comparison between Naive Influence Removal and PUMA.} (a) The projection direction of each data point (green arrow as shown Equation~\ref{eq:project_direction_single}). Blue arrow shows the one marked to remove. (b) The overall projection direction (red arrow) is toward high loss area after naive data contribution removal. (c) The overall projection direction (red arrow) is toward low loss area after PUMA data removal. Orange contour plot shows the loss surface of training objective $\mathcal{J}$. Purple contour plot shows the loss surface of performance criterion $\mathcal{C}$. For both contour plots, lighter color shows lower loss. }
    \label{fig:projection_direction}
    %\vspace{-5mm}
\end{figure*}

\subsubsection{Performance Gap as Taylor Approximation of Parameter Changes}
When the difference between two sets of parameters is reasonably small, the performance gap between the two corresponding models could be approximated through Taylor expansion such that
\begin{equation}
\begin{aligned}
        &\mathcal{C}(\theta_{\textit{mod}}) -\mathcal{C}(\theta_{\textit{org}}) = \nabla \mathcal{C}(\theta_{\textit{org}}) (\theta_{\textit{mod}} - \theta_{\textit{org}}) + \epsilon\\
        &\approx -\sum_{j=1}^{|D_{\textit{up}}|} \lambda_j \nabla \mathcal{C}(\theta_{\textit{org}})\left(\nabla^2 \mathcal{J}_{\textit{org}}(\theta_{\textit{org}})\right)^{-1}\nabla \mathcal{L}(\mathbf{x}_j, y_j, \theta_{\textit{org}}),\\
\end{aligned}
\label{eq:taylor_expansion_performance_difference}
\end{equation}
which is a linear function of the additive data perturbation $\bm{\lambda}$, where $\epsilon$ represents the higher order Taylor expansion that is exponentially smaller than the first term. Intuitively, term
\begin{equation}
    \psi(\mathbf{x}_j, y_j) =\underbrace{\nabla \mathcal{C}(\theta_{\textit{org}})\left(\nabla^2 \mathcal{J}_{\textit{org}}(\theta_{\textit{org}})\right)^{-1}}_{\text{Hessian Vector Product~(HVP)}}\nabla \mathcal{L}(\mathbf{x}_j, y_j, \theta_{\textit{org}})
\label{eq:individual_contribution}
\end{equation}
is a scalar that serves as the individual contribution score of data $(\mathbf{x}_j, y_j)$ to the performance degradation. By adjusting the weights $\bm{\lambda}$, one can control the performance gap effortlessly. Hence, at this point, we established the causal relation between data perturbation and model performance changes.

\subsection{Performance Preserved Data Removal through Gradient Re-weighting}
\label{sec:gradient_reweighting}
By combining Equation~\ref{eq:performance_gap} and Equation~\ref{eq:taylor_expansion_performance_difference}, we note they form an implicit constraint on the data up-scaling factors $\bm{\lambda}$ such that any changes on a subset factor $\lambda_{\textbf{j}}$ would encourage the changes of remaining $\bm{\lambda}_{/\textbf{j}}$ as complement to maintain the performance gap smaller than $\delta$.

Based the above notion, we describe how we remove the influence of some marked data points $D_{\textit{mk}}\subseteq D_{\textit{up}}$ from a target model $f_{\theta_{\textit{org}}}$ without hurting the model performance. 

According to the Equation~\ref{eq:optimal_modification}, removing the contribution of a marked data point $(\mathbf{x}_k,y_k)$ is equivalent to setting its perturbation factor $\lambda_k$ to $-1$. Correspondingly, to maintain the model performance while removing data points $D_{\textit{mk}}$, we propose optimizing the assignment of the perturbation factor $\bm{\lambda}$ for the remaining training data points (or randomly sampled subset $D_{\textit{up}\backslash\textit{mk}}$) to complement model criterion degradation. Concretely, we propose solving the following linear optimization task
\vspace{-3mm}
\begin{equation}
     \underset{\bm{\lambda}}{\argmin} \left\lVert\sum_{j\not\in D_{\textit{mk}} }^{|D_{\textit{up}}|} \lambda_j \psi(\mathbf{x}_j, y_j) - \sum_{k=1}^{|D_{\textit{mk}}|}\psi(\mathbf{x}_k, y_k) \right\rVert^2\!+\!\Omega(\bm{\lambda}),
\label{eq:optimization_objective}
\end{equation}
where $\Omega$ denotes the regularization term which encourages both sparsity ($l_1$ norm) and small changes of $\bm{\lambda}$ ($l_2$ norm). In terms of computational efficiency, since the $\psi(\mathbf{x}, y)$s are scalar values, the optimization is simple convex optimization.  While estimating individual contribution $\psi(\mathbf{x}_j, y_j)$ looks expensive, the estimation is no more than a dot product between individual gradient and pre-cached Hessian Vector Product (HVP) term.

With the optimized contribution factor $\bm{\lambda}^{*}$, we can then update the model parameters by a simple patching such that
\begin{equation}
        \theta_{\textit{mod}} = \theta_{\textit{org}} + \eta \left[ \sum_{k=1}^{|D_{\textit{mk}}|}\phi(\mathbf{x}_k,y_k) - \sum_{j\not\in D_{\textit{mk}} }^{|D_{\textit{up}}|} \lambda_j^{*} \phi(\mathbf{x}_j,y_j)\right],
\label{eq:puma_update}
\end{equation}
where the individual projection of each data point is
\begin{equation}
    \phi(\mathbf{x}, y) =\left(\nabla^2 \mathcal{J}_{\textit{org}}(\theta_{\textit{org}})\right)^{-1}\nabla \mathcal{L}(\mathbf{x}, y, \theta_{\textit{org}})
\label{eq:project_direction_single}
\end{equation}
and projection rate $\eta \ll 1$ is a hyper-parameter which keeps patching effective while holding our previous assumptions such that data upweighting is reasonably small.

Figure~\ref{fig:projection_direction} shows a simple example of PUMA data removal. When a data point is marked for removal (blue arrow), PUMA optimizes Equation~\ref{eq:optimization_objective} and applies the optimal factor $\bm{\lambda}$ to the projection formula (Equation~\ref{eq:project_direction_single}) to adjust model parameters such that model performance with respect to the performance criterion (purple contour) is preserved. In contrast, if we naively remove the local influence of the marked data point, the model would result in performance degradation. In this particular example, performance criterion is measured through Expected Calibration Error~(ECE)~\cite{guo2017calibration}. The example model is a linear model with two parameters trained on a binary classification task.

\section{Experiments and Evaluations}
In this section, we conduct various experiments to answer the following research questions:
\begin{itemize}[leftmargin=8pt, itemsep=0pt, partopsep=0pt, topsep=0pt]
    \item {\bf RQ1:} Is the proposed approach able to preserve model performance while removing data points? %Compare to retrain from scratch, sub-model approaches, and amnesia network.
    \item {\bf RQ2:} Is the removal successful in terms of causing membership attack failure?
    \item {\bf RQ3:} How efficient is the proposed approach compared to other state-of-the-art candidates?
    \item {\bf RQ4:} How sensitive is PUMA with respect to its hyper-parameters?
    \item {\bf RQ5:} Can the proposed approach conduct mislabeling debugging as it estimates the influence of training data point?
\end{itemize}

\subsection{Experimental Settings}

\noindent{\bf Candidate Data Removal Algorithms } 
In data removal experiments, we compare PUMA against the following state-of-the-art data removal approaches.
\begin{itemize}[leftmargin=8pt, itemsep=0pt, partopsep=0pt, topsep=0pt]
    \item {\bf Retrain Model:} Retrain model from scratch with remaining data points after picking out marked data points. 
    \item {\bf Retrain Sub-model:} Retrain sub-model that is trained on marked data points. This is also called Sharded, Isolated, Sliced, and Aggregated training~(SISA).
    \item {\bf Amnesiac Machine Learning: } Track gradient information of each training batch during training phase. Subtract the gradients when the batch is marked for removal.
\end{itemize}

\noindent {\bf Mislabelling Debugging Algorithms }
In mislabelled data debugging experiments, we compare PUMA against the following well-known debugging approaches including Influence Function~\cite{KohL17}, Representor Point Selection~\cite{yeh2018representer}, and Data Sharply Value~\cite{ghorbani2019data}.

% For both groups of the experiments, we evaluate performance as well as computation efficiency.

\noindent {\bf Datasets} We conducted our experiments on two synthetic datasets, two tabular datasets from UCI data group~\cite{Dua:2019}, and the MNIST dataset~\cite{lecun-mnisthandwrittendigit-2010}. Full description of the data used in this paper is given in Appendix A.

\begin{figure}[t]
     \centering
     \begin{subfigure}[b]{0.495\linewidth}
         \centering
         \includegraphics[width=\linewidth]{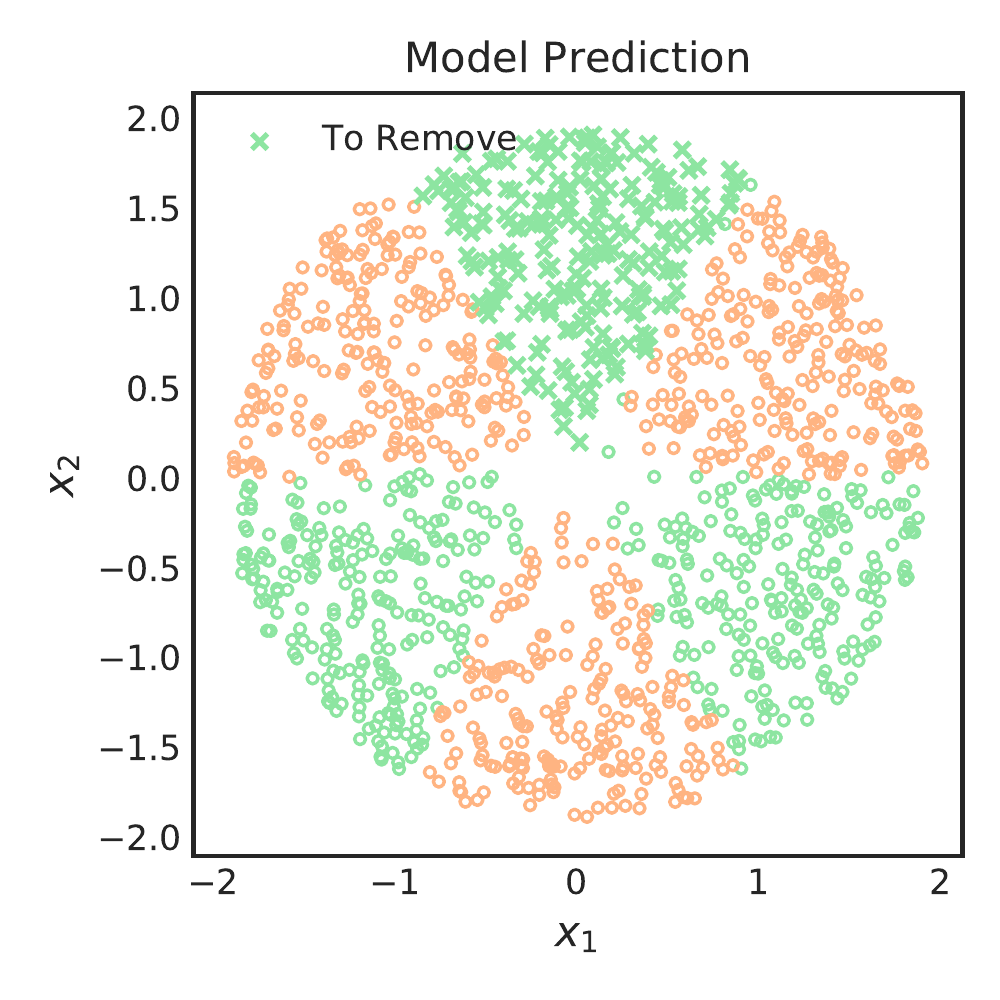}
        \caption{Before Removal}
     \end{subfigure}
     \begin{subfigure}[b]{0.495\linewidth}
        \centering
        \includegraphics[width=\linewidth]{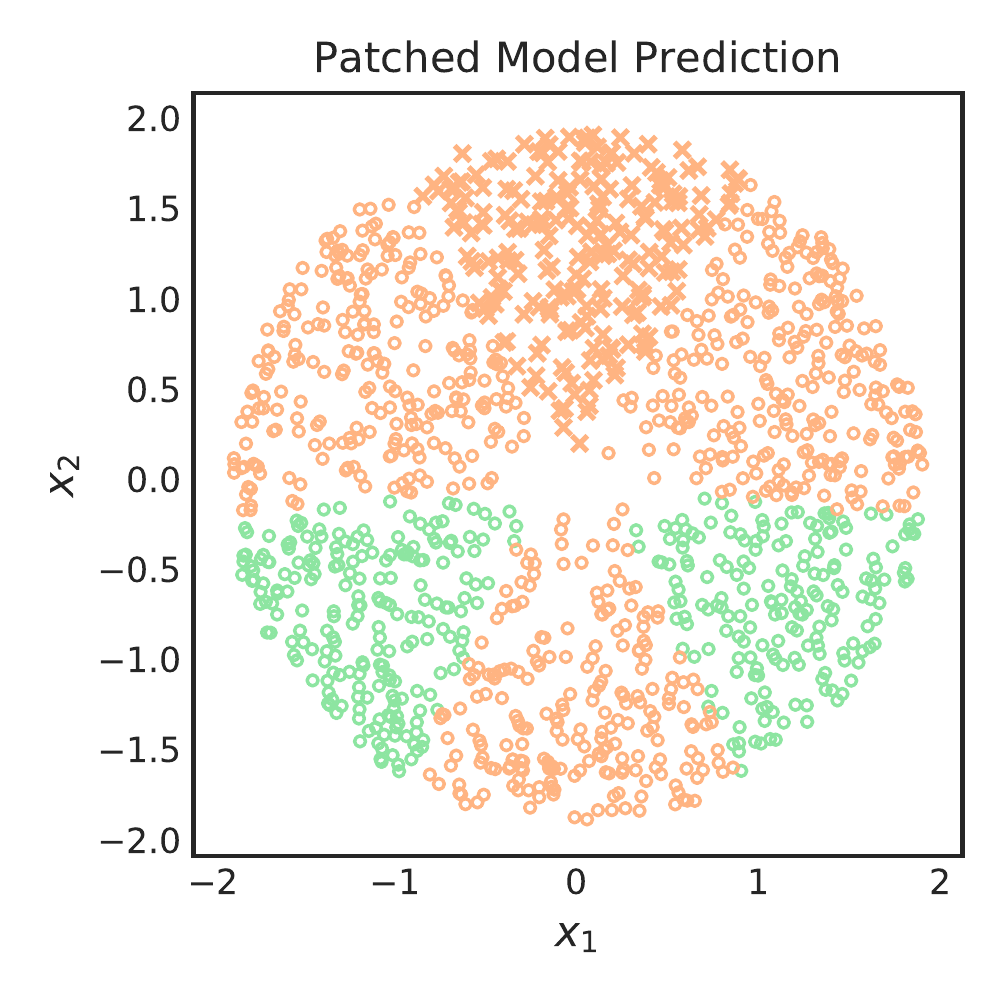}
        \caption{After Removal}
     \end{subfigure}
    \caption{Removing the training points marked by crosses from the model. As demonstrated in the right plot, PUMA successfully removed the information of all marked points. `x' in the plot shows the data intended to remove. Colors show the class labels.}
    \label{fig:demo_synthetic_radial}
\end{figure}

\begin{table*}[t]
\caption{Comparison of Model Performance Preservation among Candidate Removal Approaches. Value shows accuracy. Higher is better after data removal. We omit to present statistics in the main paper for clearness. The full table with statistics is presented in Appendix D for further reference. }
%\vspace{-3mm}
\resizebox{\linewidth}{!}{
\begin{tabular}{l|l|clcccc|clcccc}
\toprule
\multirow{2}{*}{Data Group}&\multirow{2}{*}{Dataset} &\multicolumn{6}{c|}{Ordered}&\multicolumn{6}{c}{Random}\\
\cmidrule(lr){3-14}
&& Original & Approach & 20\% & 40\% & 60\% & 80\% & Original & Approach & 20\% & 40\% & 60\% & 80\%\\
\midrule
\multirow{6}{*}{Synthetic}&\multirow{4}{*}{Radial} & 95.04 & Retrain Model & \bf{93.64} & \bf{91.60} & \bf{84.15} & 66.24 & 95.89 & Retrain Model & \bf{93.97} & \bf{90.94} & \bf{82.58} & 66.51 \\
& & 80.88 & SISA & 67.35 & 63.57 & 61.93 & 51.91 & 75.62 & SISA & 64.71 & 64.35 & 54.80 & 54.77\\
& & 95.04 & Amnesiac ML & 56.38 & 54.75 & 53.53 & 50.54 & 95.88 & Amnesiac ML & 49.08 & 48.95 & 48.95 & 48.95\\
& & 94.97 & PUMA & 68.97 & 69.60 & 67.99 & \bf{70.77} & 95.82 & PUMA & 72.44 & 73.22 & 71.82 & \bf{76.02}\\
\cmidrule(lr){2-14}
&\multirow{4}{*}{Rectangular} & 62.00 & Retrain Model & \bf{61.20} & \bf{60.35} & \bf{55.80} & 54.25 & 65.00 & Retrain Model & \bf{64.70} & \bf{64.50} & 62.30 & 58.65 \\
& & 55.60 & SISA & 55.90 & 48.30 & 30.10 & 29.55 & 56.50 & SISA & 56.50 & 56.50 & 56.55 & 56.90 \\
& & 62.00 & Amnesiac ML & 46.60 & 43.85 & 43.45 & 39.15 & 65.00 & Amnesiac ML & 35.40 & 35.40 & 35.40 & 35.40 \\
& & 61.85 & PUMA & 55.25 & 56.30 & 53.85 & \bf{61.70}  & 64.95 & PUMA & 59.90 & 62.05 & \bf{62.55} & \bf{64.80} \\
\midrule
\multirow{6}{*}{\shortstack{Tabular\\ (UCI)}}&\multirow{4}{*}{German} & 71.52 & Retrain Model & \bf{70.56} & 70.12 & 70.11 & 70.00 & 75.16 & Retrain Model & \bf{74.88} & \bf{73.24} & 72.47 & 70.00 \\
& & 70.00 & SISA & 70.00 & 70.00 & 68.96 & 66.16 & 70.00 & SISA & 70.00 & 70.00 & 70.00 & 70.00\\
& & 71.52 & Amnesiac ML & 68.52 & 64.40 & 66.24 & 64.03 & 75.16 & Amnesiac ML & 36.24 & 36.28 & 35.72 & 35.72 \\
& & 71.47 & PUMA & 69.08 & \bf{70.72} & \bf{70.64} & \bf{70.72} & 75.12 & PUMA & 70.96 & \bf{73.24} & \bf{74.44} & \bf{74.28}\\
\cmidrule(lr){2-14}
&\multirow{4}{*}{Breast Cancer} & 96.45 & Retrain Model & \bf{96.62} & \bf{96.11} & 96.00 & 94.85 & 96.00 & Retrain Model & \bf{95.82} & \bf{95.54} & \bf{95.65} & 95.20  \\
& & 91.31 & SISA & 89.20 & 88.91 & 80.68 & 52.62 & 92.28 & SISA & 91.60 & 88.05 & 88.22 & 87.88 \\
& & 96.45 & Amnesiac ML & 96.05 & 95.82 & 95.25 & 82.28 & 96.00 & Amnesiac ML & 35.20 & 30.51 & 30.51 & 30.51 \\
& & 96.39 & PUMA & 96.17 & 95.88 & \bf{96.22} & \bf{96.62} & 96.00 & PUMA & 95.08 & 94.91 & 95.25 & \bf{95.54} \\
\midrule
\multirow{3}{*}{Image}&\multirow{4}{*}{MNIST} & 97.58 & Retrain Model & \bf{97.28} & \bf{96.72} & 95.76 & 93.48 & 97.99 & Retrain Model & \textbf{97.72} & 97.16 & 96.60 & 93.98\\
& & 95.89 & SISA & 95.80 & 95.67 & 94.78 & 89.86 & 95.66 & SISA & 93.47 & 90.63 & 78.06 & 59.83\\
& & 97.44 & Amnesiac ML & 9.44 & 9.84 & 9.56 & 9.36 & 98.06 & Amnesiac ML & 10.39 & 10.39 & 10.39 & 10.39\\
& & 97.60 & PUMA & 96.70 & 96.66 & \bf{97.17} & \bf{97.16} & 97.97 & PUMA & 97.42 & \textbf{97.58} & \textbf{97.60} & \textbf{97.61}\\
\bottomrule
\end{tabular}}
\label{table:performance_perservaction}
%\vspace{-2mm}
\end{table*}

\begin{table*}[t]
\caption{Comparison of Membership Attack after Data Removal Operation. Value shows percentage of removed data that is identified as training data. Lower values in the table show better performance of removal.}
%\vspace{-3mm}
\resizebox{\linewidth}{!}{
\begin{tabular}{l|l|cc|cc|cc|cc|cc|cc|cc|cc}
\toprule
\multirow{4}{*}{\shortstack{Data\\ Group}} & \multirow{4}{*}{Dataset} & \multicolumn{8}{c|}{Ordered} & \multicolumn{8}{c}{Random} \\
\cmidrule(lr){3-18}
&&\multicolumn{2}{c|}{Retrain Model}&\multicolumn{2}{c|}{SISA}&\multicolumn{2}{c|}{Amnesiac ML}&\multicolumn{2}{c|}{PUMA}&\multicolumn{2}{c|}{Retrain Model}&\multicolumn{2}{c|}{SISA}&\multicolumn{2}{c|}{Amnesiac ML}&\multicolumn{2}{c}{PUMA}\\
\cmidrule(lr){3-18}
&&Before&After&Before&After&Before&After&Before&After&Before&After&Before&After&Before&After&Before&After\\
\midrule
\multirow{2}{*}{Synthetic}&Radial&100.00&100.00&100.00&100.00&100.00&\bf{0.00}&100.00&5.31&100.00&52.36&100.00&37.00&100.00&50.00&100.00&\bf{1.18}\\
&Rectangular&100.00&91.65&83.18&83.18&100.00&\bf{33.33}&100.00&36.66&100.00&67.07&98.50&94.00&100.00&86.20&100.00&\bf{20.00}\\
\midrule
\multirow{2}{*}{Tabular}&German&100.00&77.12&100.00&100.00&100.00&0.00&100.00&\bf{3.42}&94.44&84.44&100.00&98.81&94.44&93.33&85.18&\bf{2.22}\\
&Breast Cancer&100.00&100.00&87.50&87.50&100.00&100.00&100.00&\bf{56.25}&100.00&100.00&90.00&73.75&100.00&87.50&100.00&\bf{71.25}\\
\midrule
Image&MNIST&100.00&100.00&100.00&100.00&100.00&100.00&100.00&\bf{0.00}&100.00&100.00&100.00&100.00&100.00&100.00&100.00&\bf{72.00}\\
\bottomrule
\end{tabular}}
\label{table:membership_attack}
%\vspace{-3mm}
\end{table*}

\subsection{Dessert: Preliminary Data Removal Check}
Before starting quantitative evaluation, we first run a preliminary check on a simple binary classification task to show the effect of PUMA data removal. Specifically, we first train a classifier on a synthetic dataset that contains three observation clusters for each class as shown in Figure~\ref{fig:demo_synthetic_radial}~(a). The trained classifier is a perfect estimator of data distribution (with $100\%$ prediction accuracy). We then mark all data in one cluster for removal (denoted by `x' in the plots). Intuitively, if the marked data points are never used for training the classifier, we can imagine that their predictions should align with the predictions of data points surrounding them. Indeed, the model obtained after the PUMA data removal operation reflects our intuition as shown in Figure~\ref{fig:demo_synthetic_radial}~(b), where all removed data points are now predicted as members of the orange class.

\subsection{Effectiveness of Preserving Model Performance}
In this section, we quantitatively evaluate how the data removal approaches preserve model performance after data removal. In particular, we gradually remove training data points with percentages $[20\%, 40\%, 60\%, 80\%]$ and aim to show the performance degradation after data removal. To simplify the experimental setting, here we assume the training objective $\mathcal{J}$ and performance criterion $\mathcal{C}$ are identical (both of them are cross entropy loss of prediction). Considering that both Amnesiac ML and SISA models may show better performance when the data marked to be removed belong to same training batch, we conduct experiments in two scenarios. In the first scenario (Ordered), we intentionally group all data points marked to be removed into small set of training batches such that the removal operation would not impact other training batches (and sub-models for SISA). In the second scenario (Random), we simulate a more realistic setting where removal may apply to any data points irrespective of training batches. 

Table~\ref{table:performance_perservaction} shows performance preservation comparison between our proposed approach (PUMA) and various baselines. In the table, we make the following observations:
\begin{itemize}[leftmargin=8pt, itemsep=0pt, partopsep=0pt, topsep=0pt]
\item Among all candidate data removal approaches, PUMA shows the best performance preservation ability. And, in some cases, the model obtained after the PUMA operation even shows better performance than the original model. 
\item Amnesiac ML often completely destroys the model with its data removal operation when the removal is applied to more than 20\% of training data. This observation aligns with the original results described in the Amnesiac ML paper~\cite{Vijay2020} where refined training is required after the removal operation.
\item While Amnesiac ML and SISA show reasonably satisfactory performance preservation ability in one of the two scenarios, they tend to fail in another scenario. Amnesiac ML fails in the setting where data may be required to be removed from random batches. In contrast, SISA does not perform well when the number of sub-models is reduced, as a consequence of removing all training data points of the sub-models.
\end{itemize}

\begin{figure}[t]
     \centering
     \begin{subfigure}[b]{0.495\linewidth}
         \centering
         \includegraphics[width=\linewidth]{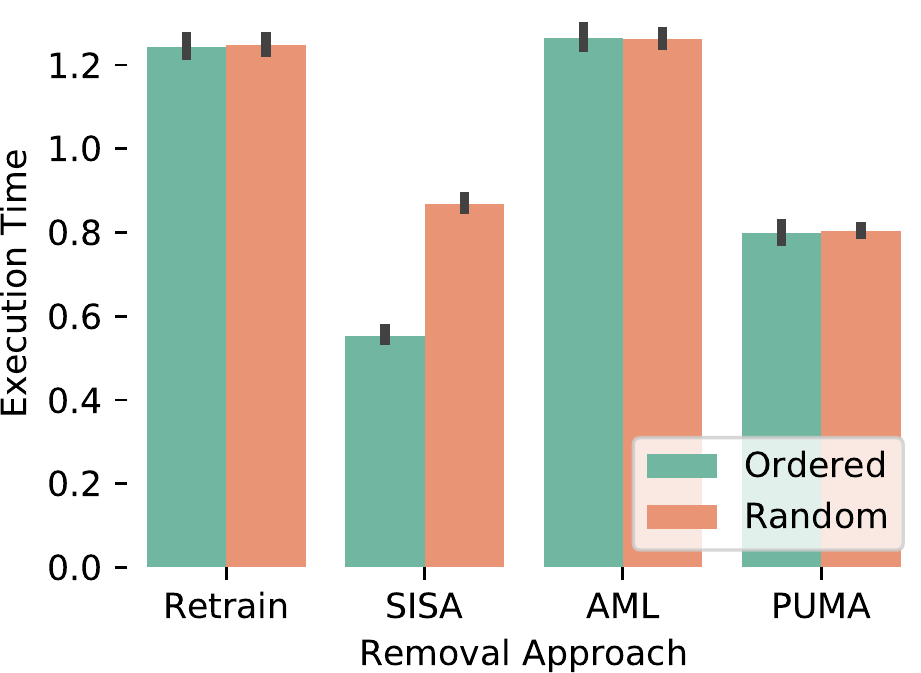}
         \caption{UCI German Credit}
     \end{subfigure}
     \begin{subfigure}[b]{0.495\linewidth}
         \centering
         \includegraphics[width=\linewidth]{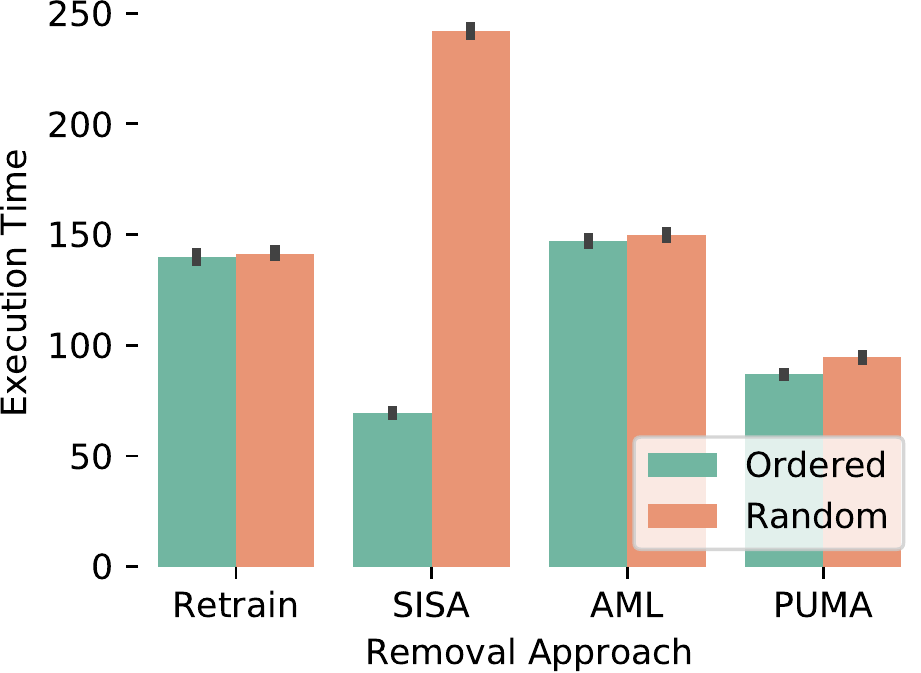}
         \caption{MNIST}
     \end{subfigure}
    %  \begin{subfigure}[b]{0.32\linewidth}
    %     \centering
    %     \includegraphics[width=0.9\linewidth]{figs/synthetic_radial_time.pdf}
    %     \caption{Image MNIST\wuga{Fake!}}
    %  \end{subfigure}
    \caption{Execution time comparison among the data removal approaches. Statistics come from 50 times run, and error bar shows the standard deviation. Lower is better.}
    \label{fig:run_time}
\end{figure}

\subsection{Effectiveness of Data Removal}
Now, we show how well the proposed approach works in removing the influence of data points from the model. To quantitatively evaluate the performance, we conduct a membership attack on the model after data removal. Ideally, if the influence of a data point is successfully removed, then the membership attack would predict that the given data point does not belong to the training data set. Hence, a lower value for data removal shows better removal effectiveness. 

Table~\ref{table:membership_attack} shows a comparison of the effectiveness of the data removal approaches. In the table, we observe follows:
\begin{itemize}[leftmargin=8pt, itemsep=0pt, partopsep=0pt, topsep=0pt]
    \item In most cases, PUMA shows better data removal performance compared to the other baseline models. While Amnesiac ML occasionally outperforms PUMA, we realize that it could be due to a complete model degradation, as previously observed in Table~\ref{table:performance_perservaction}. 
    \item In multiple experiments, we observed that the data removal operations could not reduce the success rate of membership attack to zero. This is due to the existence of similar training examples to the marked data points that are not marked for removal. Since well-train ML models can generalize well on previously unseen data points, these remaining data points can also fool the membership attack classifier when the prediction confidence is high enough.
\end{itemize}

\begin{figure*}[t]
     \centering
     \begin{subfigure}[b]{0.222\linewidth}
         \centering
         \includegraphics[width=\linewidth]{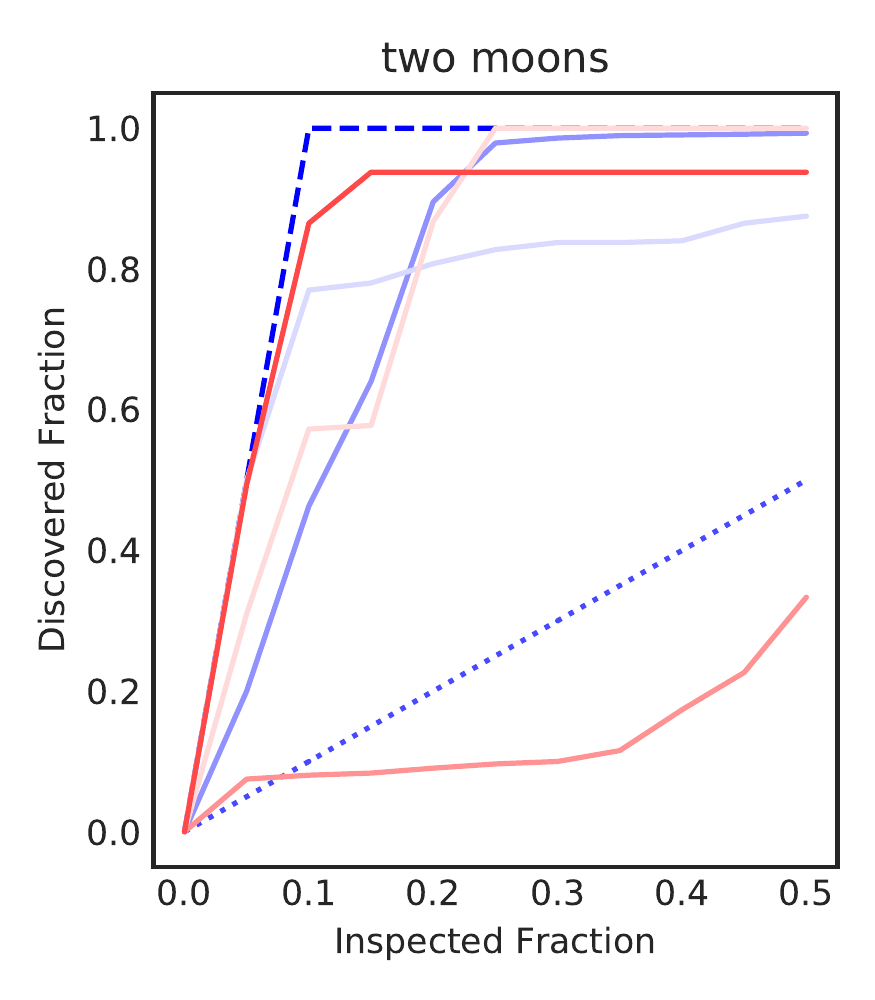}
     \end{subfigure}
     \begin{subfigure}[b]{0.222\linewidth}
         \centering
         \includegraphics[width=\linewidth]{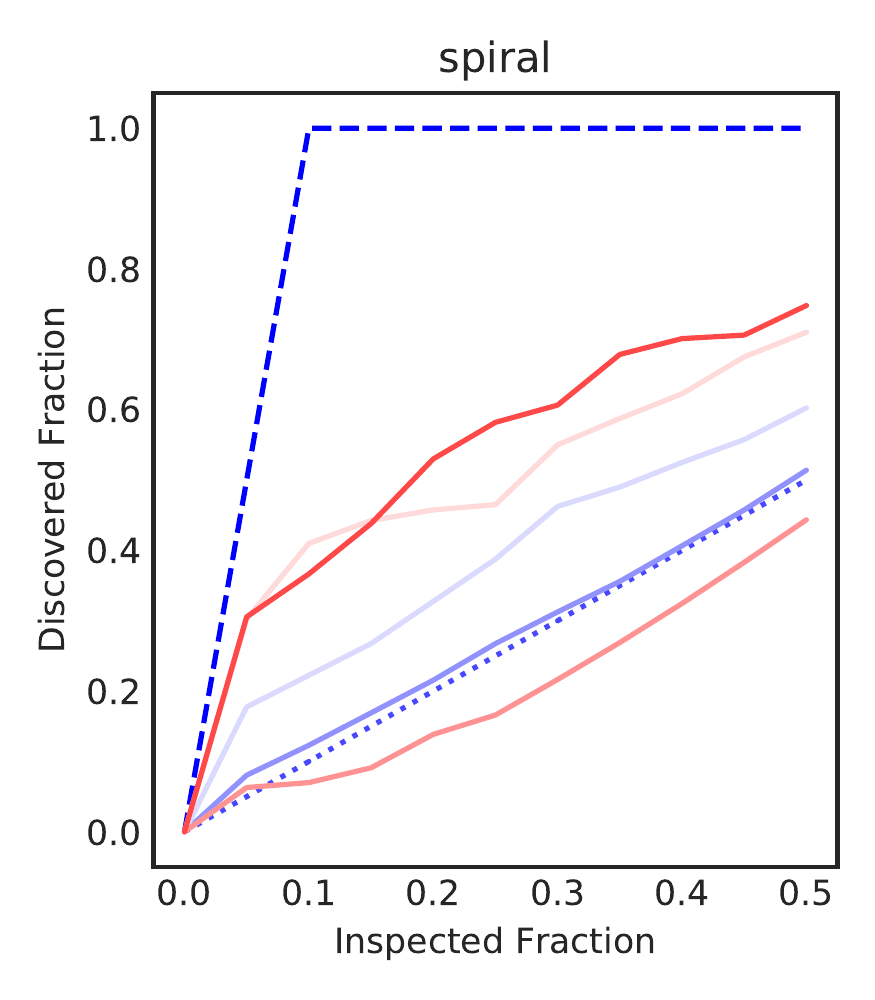}
     \end{subfigure}
     \begin{subfigure}[b]{0.222\linewidth}
        \centering
        \includegraphics[width=\linewidth]{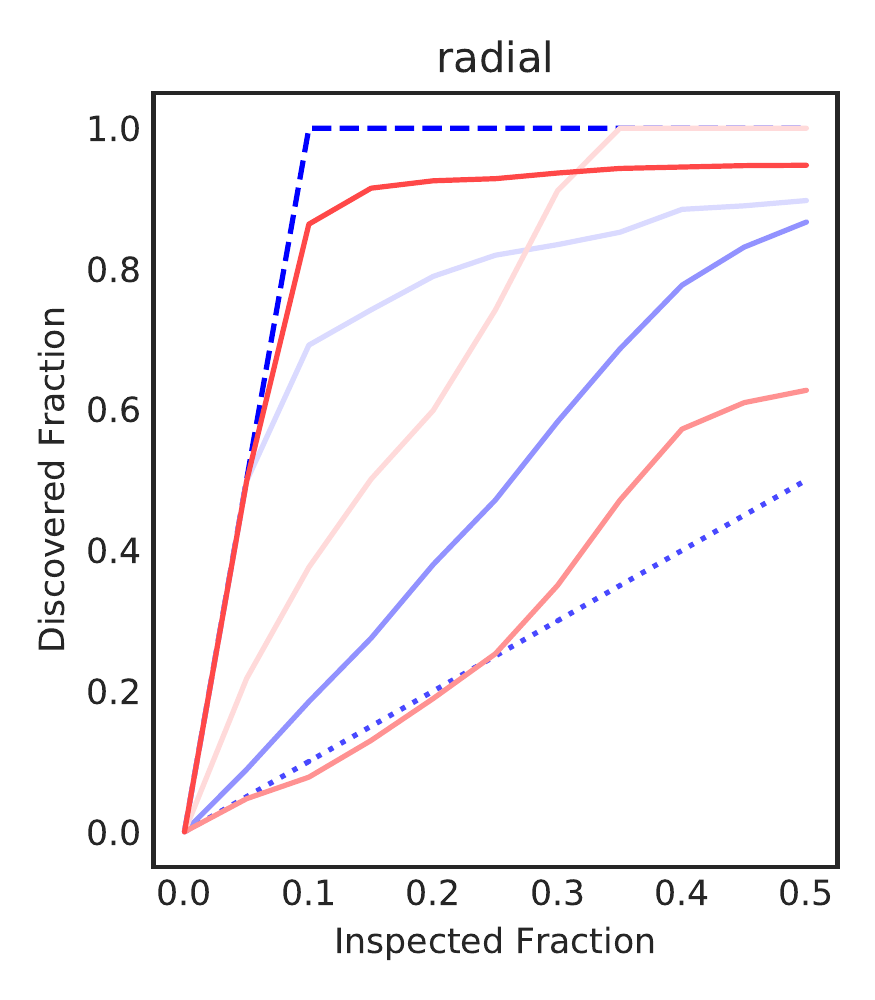}
     \end{subfigure}
     \begin{subfigure}[b]{0.317\linewidth}
        \centering
        \includegraphics[width=\linewidth]{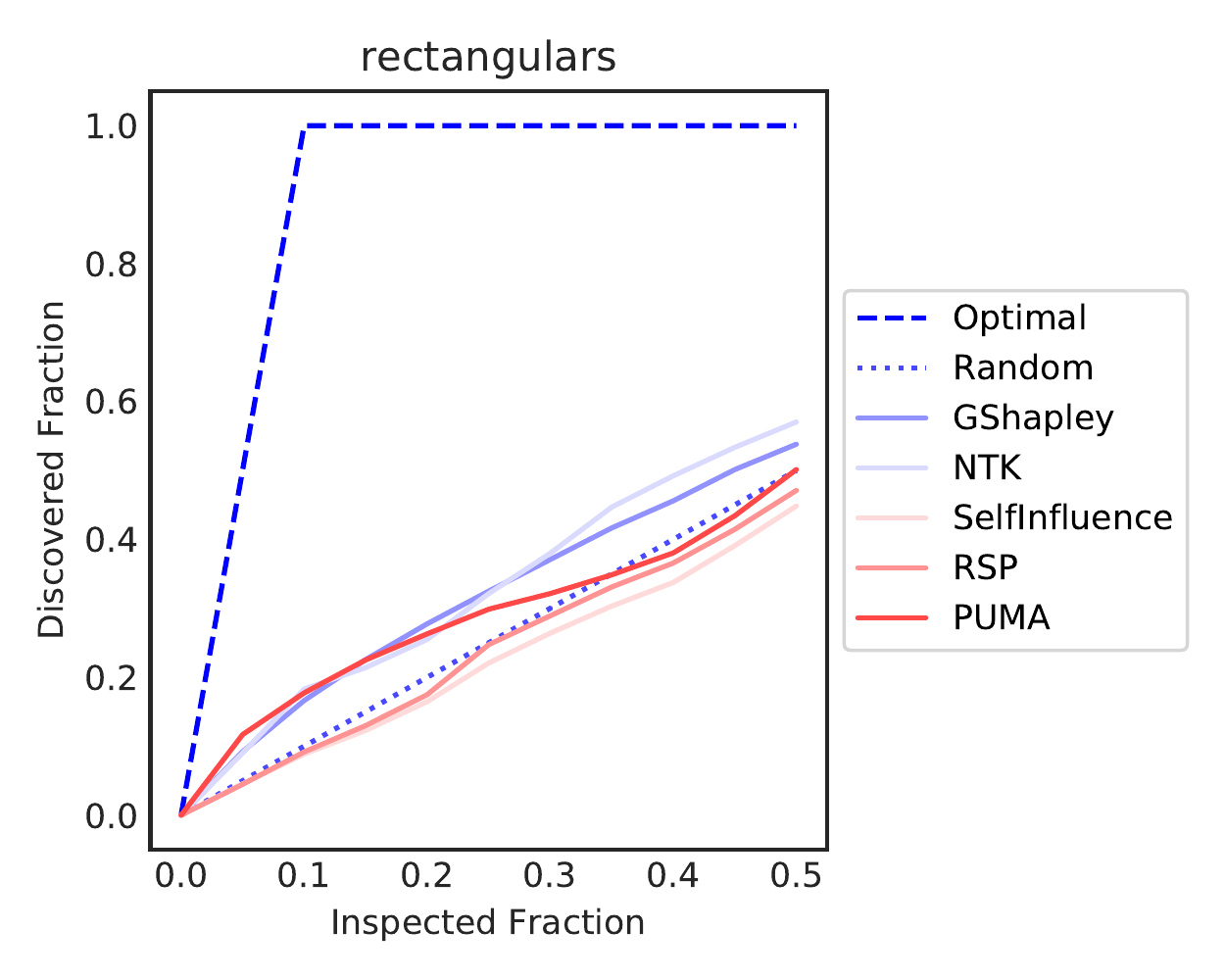}
     \end{subfigure}
    %  \begin{subfigure}[b]{0.222\linewidth}
    %     \centering
    %     \includegraphics[width=\linewidth]{figs/german_discovery.pdf}
    %  \end{subfigure}
    %  \begin{subfigure}[b]{0.222\linewidth}
    %     \centering
    %     \includegraphics[width=\linewidth]{figs/breast_w_discovery.pdf}
    %  \end{subfigure}
    \caption{Mislabelling debugging comparison between PUMA and state-of-the-art debugging algorithms. We corrupted datasets by randomly flipping 10\% of the data labels. The goal of the candidate approaches is to identify and correct the mislabelled data as early as possible. PUMA shows significant advantage when only 20\% of data are processed during debugging.}
    \label{fig:label_flip_discovery}
    %\vspace{-4mm}
\end{figure*}

% Two Moons & $90.37 \pm 2.50$ & $22.65 \pm 0.62$ & $1562.93 \pm 113.78$ & $17.14 \pm 0.17$ & $7.61 \pm 0.08$ \\
% Spiral & $78.47 \pm 0.62$ & $19.91 \pm 0.23$ & $1464.01 \pm 5.32$ & $16.51 \pm 0.13$ & $7.44 \pm 0.11$ \\
% Radial & $82.99 \pm 0.29$ & $21.60 \pm 0.42$ & $1563.53 \pm 2.54$ & $17.76 \pm 0.27$ & $7.76 \pm 0.07$ \\
% Rectangulars & $78.04 \pm 0.79$ & $20.23 \pm 0.38$ & $1480.12 \pm 9.33$ & $16.81 \pm 0.26$ & $7.29 \pm 0.08$ \\
% % German Credit & $54.30 \pm 6.13$ & $7.95 \pm 0.15$ & $687.02 \pm 14.31$ & $7.67 \pm 3.04$ & $7.35 \pm 0.27$ \\
% % Breast Cancer & $14.53 \pm 2.72$ & $3.28 \pm 0.05$ & $218.92 \pm 10.73$ & $2.42 \pm 0.07$ & $1.25 \pm 0.02$ \\

\begin{figure}[t]
     \centering
     \begin{subfigure}[b]{0.49\linewidth}
         \centering
         \includegraphics[width=0.98\linewidth]{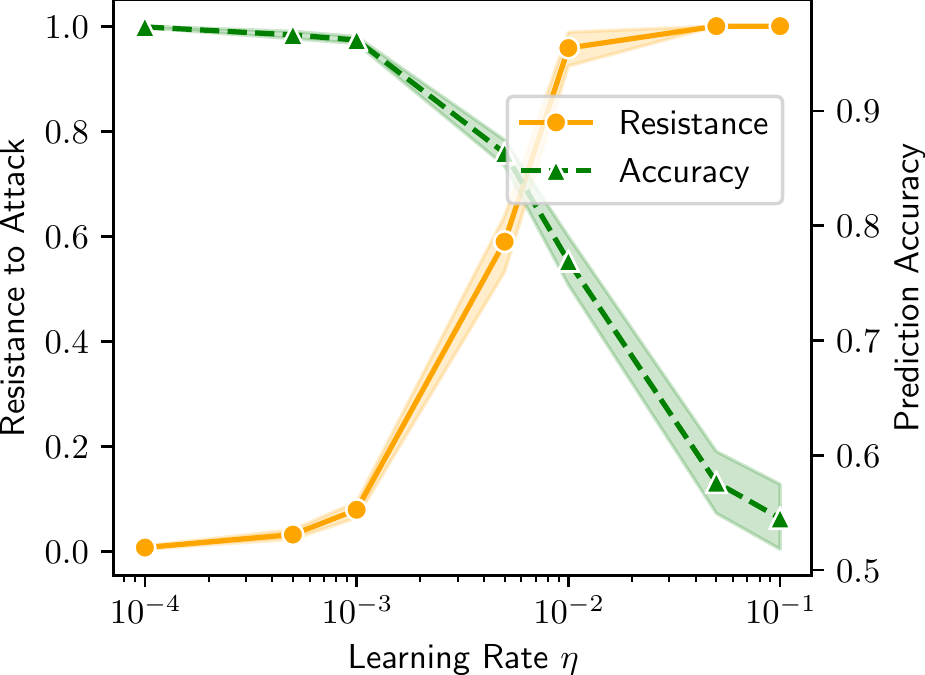}
         \caption{Synthetic Radial}
     \end{subfigure}
     \begin{subfigure}[b]{0.49\linewidth}
         \centering
         \includegraphics[width=0.98\linewidth]{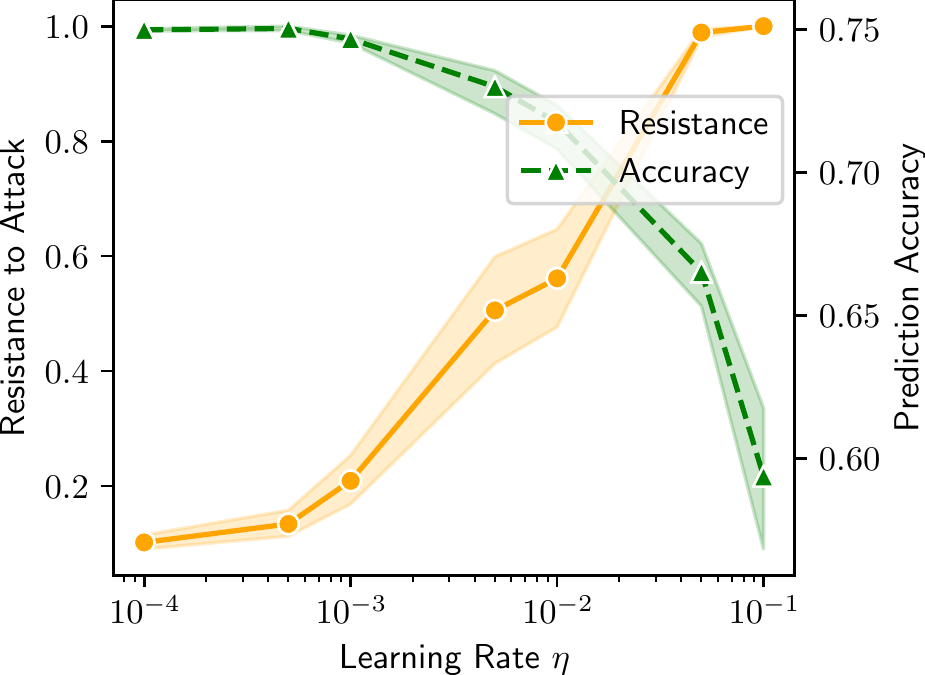}
         \caption{UCI German Credit}
     \end{subfigure}
    %  \begin{subfigure}[b]{0.32\linewidth}
    %     \centering
    %     \includegraphics[width=0.9\linewidth]{figs/synthetic_rectangular_Random.pdf}
    %     \caption{Image MNIST\wuga{Fake!}}
    %  \end{subfigure}
    \caption{Effects of hyper-parameter tuning. Large projection rate $\eta$ shows better resistant to the membership attack while suffers from severe performance degradation.}
    %\vspace{-5mm}
    \label{fig:hyper_parameter_tuning}
\end{figure}

\subsection{Efficiency of Data Removal}
As efficiency is the one of most important reason of running the data removal operation, we compare the execution time of different data removal approaches in the previously described experimental settings. Here, we only show the two most representative plots as the general trendy is similar.

Figure~\ref{fig:run_time} shows the execution time comparison on UCI German Credit and MNIST datasets. Specifically:
%The key observations are summarized as follows:
\begin{itemize}[leftmargin=8pt, itemsep=0pt, partopsep=0pt, topsep=0pt]
    \item PUMA shows the best efficiency compared to the other candidates when the data removal happens to be random (i.e. the more practical scenario). 
    \item SISA's efficiency depends on how many sub-models are involved in retraining. In the ordered data removal setting, SISA shows competitive efficiency. However, when the data removal happens to involve more sub-models, its efficiency is dramatically reduced. 
    \item In general, data removal approaches are more efficient than training a model from scratch. However, for the small dataset (UCI-German Credit), there is no significant advantage of using a data removal operation. In particular, the Amnesiac ML approach does not show better efficiency compared to retraining a model from scratch. 
\end{itemize}

\subsection{Insight of Hyper-parameter Tuning}
As introduced in Equation~\ref{eq:puma_update}, PUMA has one important hyper-parameter $\eta$ which controls the projection step of parameter augmentation. Indeed, a huge projection step $\eta$ would seriously violate the Taylor approximation assumption that PUMA approach relies on. Hence, in this experiment, we aim to demonstrate the importance of tuning this hyper-parameter.

Figure~\ref{fig:hyper_parameter_tuning} shows the trend of tuning $\eta$ on two representative datasets (UCI German Credit and MNIST~\cite{lecun-mnisthandwrittendigit-2010}). Overall, there is a trade-off between the effectiveness of removing data and the ability of preserving model generalization. Keeping the projection rate in the range of $\eta\in [10^{-2}, 10^{-1}]$ often show satisfactory removal performance while maintaining the model's generalization ability.

% \subsection{Effectiveness and Efficiency of Data Debugging}
\subsection{Corrupted Sample Discovery}
As PUMA explicitly states the contribution of individual data points to the performance criterion (see Equation~\ref{eq:individual_contribution}), a side functionality of PUMA is to debug mislabelled data in the same fashion as Influence Function~\cite{KohL17}, Representer Point Selection~\cite{yeh2018representer}, and Data Shapley~\cite{ghorbani2019data}.  
%Hence, in this experiment, we show a comparison of mislabelled data debugging among all possible candidates. Note, 
We also have included a simplified version of the Influence function by removing the inverse Hessian matrix from the influence function formulation to accelerate the computation, denoted by Neural Tangent Kernel (NTK), due to its similarity to the NTK formulation~\cite{jacot2018neural}. Figure \ref{fig:label_flip_discovery} shows the overall performance of mislabel debugging. 
%When the training data contains corrupted samples with noisy labels, an automated noisy label discovery algorithm would be highly beneficial. In an optimal discovery algorithm, the data samples with noisy labels would be detected by applying a threshold on the data values. 
In this experiment, we randomly flip the label of 10\% of the training data samples and calculate the data values using the aforementioned algorithms. 
% Figure \ref{fig:label_flip_discovery} shows the portion of discovered corrupted samples by reviewing different fractions of data starting with lowest data values. 
PUMA outperforms other algorithms by discovering more corrupted training data points while reviewing fewer data fractions. 
%In ``rectangulars'' dataset, all algorithms fail due to the complexity of the dataset. 
Tabel~\ref{table:run_time_debugging} shows the corresponding execution time for the debugging test, where we observe that PUMA is significantly more efficient than the other approaches.

\begin{table}[t]
\centering
\caption{Comparison of Running Time (in Seconds). Lower values in the table show better performance to the mislabelled data debugging. We omit the statistic in this table for saving space. Please refer to Appendix D for statistics.}
%\vspace{-2mm}
\resizebox{\linewidth}{!}{
\begin{tabular}{lccccc}
\toprule
\multirow{2}{*}{Data} & \multicolumn{5}{c}{Approach}  \\
\cmidrule(lr){2-6}
 & GShapley & NTK & SelfInfluence & RSP & PUMA \\
\midrule
Two Moons & $90.37$ & $22.65$ & $1562.93$ & $17.14$ & $\bm{7.61}$ \\
Spiral & $78.47$ & $19.91$ & $1464.01$ & $16.51$ & $\bm{7.44}$ \\
Radial & $82.99$ & $21.60$ & $1563.53$ & $17.76$ & $\bm{7.76}$ \\
Rectangulars & $78.04$ & $20.23$ & $1480.12$ & $16.81$ & $\bm{7.29}$ \\
% German Credit & $54.30 \pm 6.13$ & $7.95 \pm 0.15$ & $687.02 \pm 14.31$ & $7.67 \pm 3.04$ & $7.35 \pm 0.27$ \\
% Breast Cancer & $14.53 \pm 2.72$ & $3.28 \pm 0.05$ & $218.92 \pm 10.73$ & $2.42 \pm 0.07$ & $1.25 \pm 0.02$ \\
\bottomrule
\end{tabular}}
\label{table:run_time_debugging}
%\vspace{-5mm}
\end{table}
\section{Conclusion}
This paper presents a novel data removal approach, PUMA, which removes unique characteristics of marked training data points from a trained ML model while preserving the model's performance with respect to certain performance criterion. Compared to existing approaches which require access to the model training process, PUMA shows a significant advantage as it does not restrict how the model is trained. From various experiments, we note PUMA also demonstrates better performance compared to the baseline approaches in multiple aspects, including effectiveness, efficiency and performance preservation ability. %While this paper only includes a single-step optimization solution, we note multi-step optimization often results in better performance with additional computation cost, which would lead to our future work.

\bibliography{aaai22}

\begin{thebibliography}{26}
\providecommand{\natexlab}[1]{#1}

\bibitem[{Ateniese et~al.(2015)Ateniese, Mancini, Spognardi, Villani, Vitali,
  and Felici}]{ateniese2015hacking}
Ateniese, G.; Mancini, L.~V.; Spognardi, A.; Villani, A.; Vitali, D.; and
  Felici, G. 2015.
\newblock Hacking smart machines with smarter ones: How to extract meaningful
  data from machine learning classifiers.
\newblock \emph{International Journal of Security and Networks}, 10(3):
  137--150.

\bibitem[{Bourtoule et~al.(2019)Bourtoule, Chandrasekaran, Choquette{-}Choo,
  Jia, Travers, Zhang, Lie, and Papernot}]{Bourtoule2019}
Bourtoule, L.; Chandrasekaran, V.; Choquette{-}Choo, C.~A.; Jia, H.; Travers,
  A.; Zhang, B.; Lie, D.; and Papernot, N. 2019.
\newblock Machine Unlearning.
\newblock \emph{CoRR}, abs/1912.03817.

\bibitem[{Cauwenberghs and Poggio(2000)}]{CauwenberghsP00}
Cauwenberghs, G.; and Poggio, T.~A. 2000.
\newblock Incremental and Decremental Support Vector Machine Learning.
\newblock In Leen, T.~K.; Dietterich, T.~G.; and Tresp, V., eds.,
  \emph{Advances in Neural Information Processing Systems 13, Papers from
  Neural Information Processing Systems {(NIPS)} 2000, Denver, CO, {USA}},
  409--415. {MIT} Press.

\bibitem[{Dua and Graff(2017)}]{Dua:2019}
Dua, D.; and Graff, C. 2017.
\newblock {UCI} Machine Learning Repository.

\bibitem[{Dwork et~al.(2015)Dwork, Smith, Steinke, Ullman, and
  Vadhan}]{DworkSSUV15}
Dwork, C.; Smith, A.~D.; Steinke, T.; Ullman, J.~R.; and Vadhan, S.~P. 2015.
\newblock Robust Traceability from Trace Amounts.
\newblock In Guruswami, V., ed., \emph{{IEEE} 56th Annual Symposium on
  Foundations of Computer Science, {FOCS} 2015, Berkeley, CA, USA, 17-20
  October, 2015}, 650--669. {IEEE} Computer Society.

\bibitem[{Fredrikson, Jha, and Ristenpart(2015{\natexlab{a}})}]{FredriksonJR15}
Fredrikson, M.; Jha, S.; and Ristenpart, T. 2015{\natexlab{a}}.
\newblock Model Inversion Attacks that Exploit Confidence Information and Basic
  Countermeasures.
\newblock In Ray, I.; Li, N.; and Kruegel, C., eds., \emph{Proceedings of the
  22nd {ACM} {SIGSAC} Conference on Computer and Communications Security,
  Denver, CO, USA, October 12-16, 2015}, 1322--1333. {ACM}.

\bibitem[{Fredrikson, Jha, and
  Ristenpart(2015{\natexlab{b}})}]{fredrikson2015model}
Fredrikson, M.; Jha, S.; and Ristenpart, T. 2015{\natexlab{b}}.
\newblock Model inversion attacks that exploit confidence information and basic
  countermeasures.
\newblock In \emph{Proceedings of the 22nd ACM SIGSAC conference on computer
  and communications security}, 1322--1333.

\bibitem[{Ghorbani and Zou(2019)}]{ghorbani2019data}
Ghorbani, A.; and Zou, J. 2019.
\newblock Data shapley: Equitable valuation of data for machine learning.
\newblock In \emph{International Conference on Machine Learning}, 2242--2251.
  PMLR.

\bibitem[{Ginart et~al.(2019)Ginart, Guan, Valiant, and Zou}]{GinartGVZ19}
Ginart, A.; Guan, M.~Y.; Valiant, G.; and Zou, J. 2019.
\newblock Making {AI} Forget You: Data Deletion in Machine Learning.
\newblock In Wallach, H.~M.; Larochelle, H.; Beygelzimer, A.;
  d'Alch{\'{e}}{-}Buc, F.; Fox, E.~B.; and Garnett, R., eds., \emph{Advances in
  Neural Information Processing Systems 32, NeurIPS 2019, December 8-14, 2019,
  Vancouver, BC, Canada}, 3513--3526.

\bibitem[{Graves, Nagisetty, and Ganesh(2020)}]{Vijay2020}
Graves, L.; Nagisetty, V.; and Ganesh, V. 2020.
\newblock Amnesiac Machine Learning.
\newblock \emph{CoRR}, abs/2010.10981.

\bibitem[{Guo et~al.(2020)Guo, Goldstein, Hannun, and van~der
  Maaten}]{GuoGHM20}
Guo, C.; Goldstein, T.; Hannun, A.~Y.; and van~der Maaten, L. 2020.
\newblock Certified Data Removal from Machine Learning Models.
\newblock In \emph{Proceedings of the 37th International Conference on Machine
  Learning, {ICML} 2020, 13-18 July 2020, Virtual Event}, volume 119 of
  \emph{Proceedings of Machine Learning Research}, 3832--3842. {PMLR}.

\bibitem[{Guo et~al.(2017)Guo, Pleiss, Sun, and
  Weinberger}]{guo2017calibration}
Guo, C.; Pleiss, G.; Sun, Y.; and Weinberger, K.~Q. 2017.
\newblock On calibration of modern neural networks.
\newblock In \emph{International Conference on Machine Learning}, 1321--1330.
  PMLR.

\bibitem[{Hitaj, Ateniese, and Perez-Cruz(2017)}]{hitaj2017deep}
Hitaj, B.; Ateniese, G.; and Perez-Cruz, F. 2017.
\newblock Deep models under the GAN: information leakage from collaborative
  deep learning.
\newblock In \emph{Proceedings of the 2017 ACM SIGSAC Conference on Computer
  and Communications Security}, 603--618.

\bibitem[{Homer et~al.(2008)Homer, Szelinger, Redman, Duggan, Tembe, Muehling,
  Pearson, Stephan, Nelson, and Craig}]{homer2008resolving}
Homer, N.; Szelinger, S.; Redman, M.; Duggan, D.; Tembe, W.; Muehling, J.;
  Pearson, J.~V.; Stephan, D.~A.; Nelson, S.~F.; and Craig, D.~W. 2008.
\newblock Resolving individuals contributing trace amounts of DNA to highly
  complex mixtures using high-density SNP genotyping microarrays.
\newblock \emph{PLoS Genet}, 4(8): e1000167.

\bibitem[{Huang et~al.(2017)Huang, Liu, Van Der~Maaten, and
  Weinberger}]{huang2017densely}
Huang, G.; Liu, Z.; Van Der~Maaten, L.; and Weinberger, K.~Q. 2017.
\newblock Densely connected convolutional networks.
\newblock In \emph{Proceedings of the IEEE conference on computer vision and
  pattern recognition}, 4700--4708.

\bibitem[{Jacot, Gabriel, and Hongler(2018)}]{jacot2018neural}
Jacot, A.; Gabriel, F.; and Hongler, C. 2018.
\newblock Neural tangent kernel: Convergence and generalization in neural
  networks.
\newblock \emph{arXiv preprint arXiv:1806.07572}.

\bibitem[{Karasuyama and Takeuchi(2009)}]{KarasuyamaT09}
Karasuyama, M.; and Takeuchi, I. 2009.
\newblock Multiple incremental decremental learning of support vector machines.
\newblock \emph{Advances in neural information processing systems}, 22:
  907--915.

\bibitem[{Koh and Liang(2017)}]{KohL17}
Koh, P.~W.; and Liang, P. 2017.
\newblock Understanding Black-box Predictions via Influence Functions.
\newblock In Precup, D.; and Teh, Y.~W., eds., \emph{Proceedings of the 34th
  International Conference on Machine Learning, {ICML} 2017, Sydney, NSW,
  Australia, 6-11 August 2017}, volume~70 of \emph{Proceedings of Machine
  Learning Research}, 1885--1894. {PMLR}.

\bibitem[{LeCun and Cortes(2010)}]{lecun-mnisthandwrittendigit-2010}
LeCun, Y.; and Cortes, C. 2010.
\newblock {MNIST} handwritten digit database.

\bibitem[{Lyu and Chen(2021)}]{lyu2021novel}
Lyu, L.; and Chen, C. 2021.
\newblock A Novel Attribute Reconstruction Attack in Federated Learning.
\newblock \emph{arXiv preprint arXiv:2108.06910}.

\bibitem[{Nasr, Shokri, and Houmansadr(2018)}]{Milad18}
Nasr, M.; Shokri, R.; and Houmansadr, A. 2018.
\newblock Comprehensive Privacy Analysis of Deep Learning: Stand-alone and
  Federated Learning under Passive and Active White-box Inference Attacks.
\newblock \emph{CoRR}, abs/1812.00910.

\bibitem[{Shokri et~al.(2017)Shokri, Stronati, Song, and
  Shmatikov}]{shokri2017membership}
Shokri, R.; Stronati, M.; Song, C.; and Shmatikov, V. 2017.
\newblock Membership inference attacks against machine learning models.
\newblock In \emph{2017 IEEE Symposium on Security and Privacy (SP)}, 3--18.
  IEEE.

\bibitem[{Tsai, Lin, and Lin(2014)}]{TsaiLL14}
Tsai, C.; Lin, C.; and Lin, C. 2014.
\newblock Incremental and decremental training for linear classification.
\newblock In Macskassy, S.~A.; Perlich, C.; Leskovec, J.; Wang, W.; and Ghani,
  R., eds., \emph{The 20th {ACM} {SIGKDD} International Conference on Knowledge
  Discovery and Data Mining, {KDD} '14, New York, NY, {USA} - August 24 - 27,
  2014}, 343--352. {ACM}.

\bibitem[{Yang et~al.(2021)Yang, Zhong, Liu, Wang, Luo, Li, Sebe, and
  Satoh}]{yang2021learning}
Yang, F.; Zhong, Z.; Liu, H.; Wang, Z.; Luo, Z.; Li, S.; Sebe, N.; and Satoh,
  S. 2021.
\newblock Learning to Attack Real-World Models for Person Re-identification via
  Virtual-Guided Meta-Learning.
\newblock In \emph{Proceedings of the AAAI Conference on Artificial
  Intelligence}, volume~35, 3128--3135.

\bibitem[{Yeh et~al.(2018)Yeh, Kim, Yen, and Ravikumar}]{yeh2018representer}
Yeh, C.-K.; Kim, J.~S.; Yen, I.~E.; and Ravikumar, P. 2018.
\newblock Representer point selection for explaining deep neural networks.
\newblock \emph{arXiv preprint arXiv:1811.09720}.

\bibitem[{Yeom et~al.(2018)Yeom, Giacomelli, Fredrikson, and Jha}]{YeomGFJ18}
Yeom, S.; Giacomelli, I.; Fredrikson, M.; and Jha, S. 2018.
\newblock Privacy Risk in Machine Learning: Analyzing the Connection to
  Overfitting.
\newblock In \emph{31st {IEEE} Computer Security Foundations Symposium, {CSF}
  2018, Oxford, United Kingdom, July 9-12, 2018}, 268--282. {IEEE} Computer
  Society.

\end{thebibliography}
\clearpage
\section{Appendix A: Detailed Experiment Setup}
\subsection{Dataset}
The experiments in this paper involves three groups of data sources, including synthetically generated data, UCI data~\cite{Dua:2019}, and the MNIST dataset~\cite{lecun-mnisthandwrittendigit-2010}. Table~\ref{table:dataset_summary} summarizes the datasets used in this paper.

\begin{table}[h]
\centering
\caption{Dataset used in the experiments.}
\resizebox{\linewidth}{!}{
\begin{tabular}{lcccl}
\toprule
Dataset & Size & Num of Features & Num of Labels & Model Architecture  \\
\midrule
Radial & 600 & 2 & 2 & 2-layer FCN\\
Rectangular & 800 & 2 & 3 & 2-layer FCN\\
\midrule
Breast Cancer & 699 & 9 & 2 & 2-layer FCN\\
German Credit & 1000 & 20 & 2 & 2-layer FCN\\
MNIST & 70000 & 784 & 10 & DenseNet\\
\bottomrule
\end{tabular}}
\label{table:dataset_summary}
\end{table}

\subsubsection{Synthetic Data}
\begin{figure}[h]
     \centering
     \begin{subfigure}[]{0.46\linewidth}
         \centering
         \includegraphics[width=\linewidth]{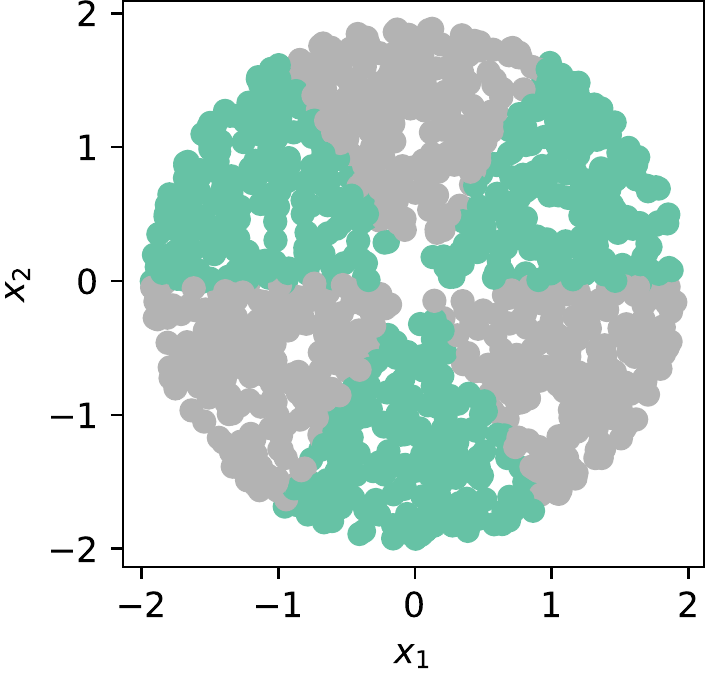}
        \caption{Radial}
     \end{subfigure}
     \begin{subfigure}[]{0.49\linewidth}
        \centering
        \includegraphics[width=\linewidth]{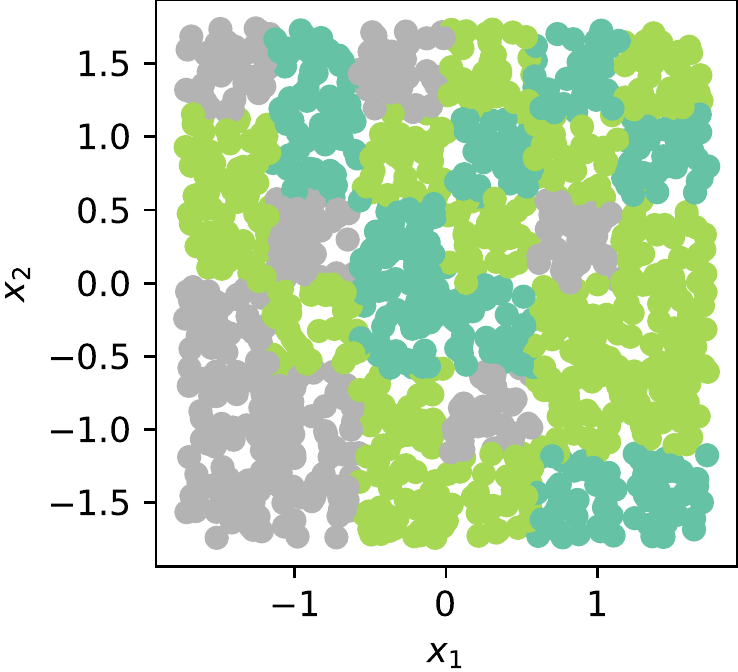}
        \caption{Rectangular}
     \end{subfigure}
    \caption{Synthetic data used in our experiments. Radial dataset has 2 classes with 6 clusters, while rectangular dataset has 3 classes with 16 clusters.}
    \label{fig:synthetic_data_train}
\end{figure}

We used multiple synthetic datasets to conduct the proof of concept experiments in the paper as the results are easy to visualize and interpret. Figure~\ref{fig:synthetic_data_train} shows the synthetic data we produced for our experiments. 

\subsubsection{UCI datasets} The UCI data is a group of classic tabular datasets well suited to justify the quantitative performance of a ML model in the literature. Among the hundreds of datasets, we chose to use German Credit and Breast Cancer dataset in our experiments since both of the datasets are closely related to two important machine learning application fields: finance and medicine. 

\subsubsection{MNIST} We include the MNIST dataset along with the Convolution Network (DenseNet) architecture in our experiments to show that the proposed approach can work on more complex model architectures than simple Fully Connected Networks~(FNC). In addition to the quantitative evaluations of the main paper, we also use it to visualize the effects of data removal in Appendix C.

\subsection{Predictive Models}
In the experiments, we train the classifiers with two neural network architectures. Specifically, for the synthetic and the tabular (UCI) data, we train fully connected neural networks that have two hidden layers with dimension of 64 and 32 respectively. For the MNIST dataset, we train a DenseNet model~\cite{huang2017densely} to reach a better prediction performance.

The following pytorch code shows the actual architecture we used in the experiments.

\begin{lstlisting}[language=Python, caption=Fully Connect Network, basicstyle=\tiny]
class FlexibleFCN(nn.Module):
    def __init__(self, neurons, activation):
        super(FlexibleFCN, self).__init__()
        self.layers = nn.ModuleList()
        for i in range(len(neurons) - 1):
            self.layers.append(torch.nn.Linear(neurons[i], neurons[i + 1]))
        if activation == "relu":
            self.activation = torch.nn.ReLU()
        else:
            self.activation = torch.nn.Sigmoid()

    def forward(self, x):
        for i in range(len(self.layers) - 1):
            x = self.layers[i](x)
            x = self.activation(x)
        x = self.layers[-1](x)
        return x
\end{lstlisting}

\begin{lstlisting}[language=Python, caption=CNN DenseNet, basicstyle=\tiny]
class DenseNet(nn.Module):
    def __init__(
        self,
        block,
        nblocks,
        growth_rate=12,
        reduction=0.5,
        num_classes=10,
        channel=3,
        last_pool_size=4,
    ):
        super(DenseNet, self).__init__()
        self.growth_rate = growth_rate

        num_planes = 2 * growth_rate
        self.conv1 = nn.Conv2d(
            channel, num_planes, kernel_size=3, 
            padding=1, bias=False
        )

        self.dense1 = self._make_dense_layers(block, 
                                              num_planes, 
                                              nblocks[0])
        num_planes += nblocks[0] * growth_rate
        out_planes = int(math.floor(num_planes * reduction))
        self.trans1 = Transition(num_planes, out_planes)
        num_planes = out_planes

        self.dense2 = self._make_dense_layers(block, 
                                              num_planes, 
                                              nblocks[1])
        num_planes += nblocks[1] * growth_rate
        out_planes = int(math.floor(num_planes * reduction))
        self.trans2 = Transition(num_planes, out_planes)
        num_planes = out_planes

        self.dense3 = self._make_dense_layers(block, 
                                              num_planes, 
                                              nblocks[2])
        num_planes += nblocks[2] * growth_rate
        out_planes = int(math.floor(num_planes * reduction))
        self.trans3 = Transition(num_planes, out_planes)
        num_planes = out_planes

        self.dense4 = self._make_dense_layers(block, 
                                              num_planes, 
                                              nblocks[3])
        num_planes += nblocks[3] * growth_rate

        self.bn = nn.BatchNorm2d(num_planes)
        self.linear = nn.Linear(num_planes, num_classes)
        self.softmax = torch.nn.Softmax(dim=1)
        self.num_classes = num_classes

        self.last_pool_size = last_pool_size

    def _make_dense_layers(self, block, in_planes, nblock):
        layers = []
        for i in range(nblock):
            layers.append(block(in_planes, self.growth_rate))
            in_planes += self.growth_rate
        return nn.Sequential(*layers)

    def forward(self, x):
        out = self.conv1(x)
        out = self.trans1(self.dense1(out))
        out = self.trans2(self.dense2(out))
        out = self.trans3(self.dense3(out))
        out = self.dense4(out)
        out = F.avg_pool2d(F.relu(self.bn(out)), 
                           self.last_pool_size)
        out = out.view(out.size(0), -1)
        out = self.linear(out)
        return out
\end{lstlisting}
\subsection{Membership Attack Setup}
We implemented the attack model based on the description in the original membership attack paper~\cite{shokri2017membership}. Since, in our experimental setting, the training data is fully observable, we use the training data to train the shadow models. Here, the shadow models hold the same architecture as the predictive models described previously. 

We train a membership attack model for each of the datasets using five shadow models, where each shadow model is trained on a subset of data points (less than 10\% of training set) randomly sampled from the training set. The number of training epochs for the shadow models is set to 50. 

After training the shadow models, a neural network based binary classifier is trained on the output of the shadow models to predict whether a data point has been a member of the training set or not. The architecture of this classifier is a two-hidden layer neural network with latent dimensions of 128 and 64. This architecture aligns with the description in the previous data removal literature~\cite{Vijay2020}. The binary classifier is then used to attack the predictive models before and after data removal to demonstrate the effectiveness of the data removal algorithms.

\subsection{Data Marking for Removal Experiment}
As mentioned in the main paper, when many similar data points exist in a dataset, removing one or a small set of them does not help to demonstrate the data removal performance (via membership attack) since the ML model would generalize to the removed data points with high prediction confidence. 

Hence, we mark the data through a clustering based approach. Specifically, we conduct the following steps to mark the data for removal experiments:
\begin{enumerate}
    \item For each class $y$ in the training set, run a k-means clustering algorithm on the features $\mathbf{X}$.
    \item Choose all data points in a cluster from the k-means clusters as the marked data points.
    \item Run step 1-2 for all classes.
\end{enumerate}

The following python code shows how we did it in our implementation.
%Note, in some demonstrative experiments, we disabled the clustering based approach and marked all data points in a randomly selected class for removal. We explicitly stated where we disabled the clustering in the paper. However, by default, all experiments are conducted with the clustering marking approach.
\begin{lstlisting}[language=Python, caption=Mark Data to Remove, basicstyle=\tiny]
@staticmethod
def mark_data(data_loader, label_interested, clustering=True):
    with torch.no_grad():
        tensor_X, tensor_y = data_loader.dataset.tensors
        label_index = torch.where(tensor_y == label_interested)[0].cpu().numpy()
        if not clustering:
            data_index_to_remove = label_index
        else:
            x = tensor_X.cpu().numpy()[label_index]
            kmeans = KMeans(n_clusters=8, random_state=0).fit(
                np.array(x).reshape(len(x), -1)
            )
            values, counts = np.unique(kmeans.labels_, return_counts=True)
            value_pri = values[counts.argsort()[0]]
            data_index_to_remove = label_index[np.where(kmeans.labels_ == value_pri)]
    return data_index_to_remove
\end{lstlisting}

\subsection{How does Amnesiac ML and SISA Handle Single/Batch Data Removal?}
Amnesiac ML does not support single data removal operation since it requires storing the gradient information for each point, which is barely possible in practice. Hence, to remove contribution of a single data point, we need to remove the entire training batch containing the point in question. 

Indeed, in the Experiment and Evaluation section, we mentioned that we have two experiment scenarios, namely Ordered and Random. In the Ordered scenario, we assume all data points that are marked to be removed belong to the same data batch. In this particular setting, Amnesiac ML could work as it was presented in its original paper~\cite{Vijay2020}. We introduce this scenario for the purpose of fair comparison. For a more realistic scenario (Random), since the marked data points may be spread across many batches, we have no choice but to remove the batches that contain the marked data points. Hence, the updated model could face severe performance degradation, as shown in Table~\ref{table:performance_perservaction}.

SISA naturally supports removing single a data point since it retrains the sub-model that has been trained on the marked data point. However, SISA faces a computational challenge when we remove multiple data points requiring to retrain multiple involved sub-models simultaneously. Hence, SISA's efficiency depends on the distribution of marked data points in the sub-model training datasets. This intuition could be justified in our experiment results (Figure~\ref{fig:run_time}), where we show SISA's run time varies when we switch scenarios (Random vs. Ordered).

\section{Appendix B: Algorithm Implementation}
In this section, we present the implementation details of the PUMA algorithm in the form of pseudo-code. 
\subsection{PUMA Removal Pseudo Code}
Algorithm~\ref{alg:puma_removal} shows the pseudo-code of the PUMA data removal procedure.  Here, we use $D_{\textit{up}}$ instead of $D_{\textit{up}/\textit{mk}}$ to represent the reweight data points for simplicity.
\vspace{-2mm}
\begin{algorithm}[h]
\caption{PUMA Data Characteristics Removal}\label{alg:puma_removal}
\begin{algorithmic}
\Require Original model $\theta_{\textit{org}}$, Training objective $\mathcal{J}_{\textit{org}}$, Performance criterion $\mathcal{C}$, Data marked to remove $D_{\textit{mk}}$, Whole training set $D_{\textit{tn}}$, Learning rate $\eta$
\Statex{\bf Optional:} Upweight data $D_{\textit{up}}$ 
\Ensure $\mathcal{C}$ is differentiable, $D_{\textit{mk}}\subset D_{\textit{tn}}$
\State \tikzmark{left} 
\State $\nabla \mathcal{C} \gets \textit{Gradient}(\mathcal{C}, D_{\textit{tn}}, \theta_{\textit{org}})$ 
\State $\nabla \mathcal{C}\left(\nabla^2 \mathcal{J}_{\textit{org}}\right)^{-1}\!\!\gets \textit{HVP}(\mathcal{J}_{\textit{org}}, D_{\textit{tn}}, \theta_{\textit{org}},  \nabla \mathcal{C})$ \Comment{Cache}
\State $D_{\textit{up}} \gets \textit{Sample}(D_{\textit{tn}}, \textit{excluding}=D_{\textit{mk}})$ \Comment{Optional}
\State\tikzmark{top1}
\For{$(\mathbf{x}_k, y_k)\in D_{\textit{mk}}$}  \Comment{Get Data Influence}
    \State $\psi(\mathbf{x}_k, y_k)\!\gets\!\nabla \mathcal{C}\!\left(\nabla^2 \mathcal{J}_{\textit{org}}\right)^{\!-1}\!\nabla \mathcal{L}(\mathbf{x}_k, y_k, \theta_{\textit{org}})$
\EndFor 
\State
\For{$(\mathbf{x}_j, y_j)\in D_{\textit{up}}$} \Comment{Get Data Influence}
    \State $\psi(\mathbf{x}_j, y_j)\!\gets\!\underbrace{\nabla \mathcal{C}\!\left(\nabla^2 \mathcal{J}_{\textit{org}}\right)^{\!-1}}_{\textit{Cached}}\!\nabla \mathcal{L}(\mathbf{x}_j, y_j, \theta_{\textit{org}})$
\tikzmark{bottom1}\EndFor 
\State
\State $\bm{\lambda}^*\!\gets\!\underset{\bm{\lambda}}{\argmin} \left\lVert\sum_{j} \lambda_j \psi(\mathbf{x}_j, y_j)\!-\!\sum_{k}\psi(\mathbf{x}_k, y_k) \right\rVert^2\!+\!\Omega(\bm{\lambda})$ \Comment{SLSQP Constraint Optimization}
\State\tikzmark{top2}
\State $V_{\textit{mk}} \gets 0$ \Comment{Get Parameter Projection Direction}
\For{$(\mathbf{x}_k, y_k)\in D_{\textit{mk}}$}
    \State $V_{\textit{mk}}\gets V_{\textit{mk}} +\nabla \mathcal{L}(\mathbf{x}_k, y_k, \theta_{\textit{org}})$
\EndFor 
\State $\Phi_{\textit{mk}}\gets \textit{HVP}(\mathcal{J}_{\textit{org}}, D_{\textit{tn}}, \theta_{\textit{org}},  V_{\textit{mk}})$ 
\State
\State $V_{\textit{up}} \gets 0$ \Comment{Get Parameter Projection Direction}
\For{$(\mathbf{x}_j, y_j)\in D_{\textit{mk}}$}
    \State $V_{\textit{up}}\gets V_{\textit{up}} +\lambda_j \nabla \mathcal{L}(\mathbf{x}_j, y_j, \theta_{\textit{org}})$
\EndFor 
\State $\Phi_{\textit{up}}\gets \textit{HVP}(\mathcal{J}_{\textit{org}}, D_{\textit{tn}}, \theta_{\textit{org}},  V_{\textit{up}})$ \tikzmark{bottom2}
\State
\State $\theta_{\textit{mod}} \gets \theta_{\textit{org}} + \eta \left[\Phi_{\textit{mk}} -  \Phi_{\textit{up}}\right]$ \Comment{Update Model}
\end{algorithmic}
\AddNote{top1}{bottom1}{left}{}
\AddNote{top2}{bottom2}{left}{}
\vspace{-2mm}
\end{algorithm}
\vspace{-2mm}

Overall, there are five fundamental steps in the procedure.
\begin{enumerate}
    \item Compute and cache the Hessian Vector Product $\nabla \mathcal{C}\left(\nabla^2 \mathcal{J}_{\textit{org}}\right)^{-1}$ for both the training objective $\mathcal{J}$ and performance criterion $\mathcal{C}$ on training data $D_{\textit{tn}}$.
    \item Estimate the influence value of the data points marked for removal $D_{\textit{mk}}$ and the reweighted data points $D_{\textit{up}}$.
    \item Optimize the reweighting weights $\bm{\lambda}$ with constrained optimization algorithms. E.g. SLSQP or L-BFGS.
    \item Estimate the weighted parameter projection directions for both $D_{\textit{mk}}$ and $D_{\textit{up}}$
    \item Update the model parameters with learning rate $\eta$
\end{enumerate}
While computing the inverse of the Hessian matrix is possible for simple linear models, it is generally infeasible for more complicated setups, since there is no guarantee on the Hessian matrix to be positive definite, as noticed in the previous literature~\cite{KohL17}. Hence, in our implementation, we directly compute the Hessian Vector Product approximation to avoid the potential numerical issue.

\subsection{PUMA Data Debugging Pesudo Code}
While PUMA was originally not designed for debugging mislabeled data, we note it shows reasonably good performance (with significant advantage in efficiency) compared with the state-of-the-art approaches. 

As mentioned in the main paper, since PUMA computes the influence of each data point on the overall model performance (see Equation~\ref{eq:individual_contribution}), we can use this information to rank the data points and identify the mislabelled data points that usually have large negative influences. In addition, since the data points close to the decision boundary are often noisy, we filter the candidate data points with this condition to avoid large false positive predictions. Algorithm~\ref{alg:puma_debugging} shows the algorithm for debugging mislabelled data.
\vspace{-2mm}
\begin{algorithm}[h]
\caption{PUMA Problematic Data Debugging}\label{alg:puma_debugging}
\begin{algorithmic}
\Require Original model $\theta_{\textit{org}}$, Training objective $\mathcal{J}_{\textit{org}}$, Performance criterion $\mathcal{C}$, Whole training set $D_{\textit{tn}}$, Number of problematic data to return $k$
\Ensure $\mathcal{C}$ is differentiable
\State \tikzmark{left} 
\State $\nabla \mathcal{C} \gets \textit{Gradient}(\mathcal{C}, D_{\textit{tn}}, \theta_{\textit{org}})$
\State $\nabla \mathcal{C}\left(\nabla^2 \mathcal{J}_{\textit{org}}\right)^{-1}\!\!\gets \textit{HVP}(\mathcal{J}_{\textit{org}}, D_{\textit{tn}}, \theta_{\textit{org}},  \nabla \mathcal{C})$ \Comment{Cache}
\State\tikzmark{top1}
\State $\textit{Confidence}\gets \textit{EmptyList}()$ 
\State $\textit{Influence}\gets \textit{EmptyList}()$ 
\For{$(\mathbf{x}_i, y_i)\in D_{\textit{tn}}$} 
    \State $\textit{Confidence}[i]\gets p(y_i|\mathbf{x}_i;\theta_{\textit{org}})$ \Comment{Prediction}
    \State $\textit{Influence}[i]\gets\underbrace{\nabla\mathcal{C}\!\left(\nabla^2 \mathcal{J}_{\textit{org}}\right)^{\!-1}}_{\textit{Cached}}\!\nabla \mathcal{L}(\mathbf{x}_i, y_i, \theta_{\textit{org}})$
\tikzmark{bottom1}\EndFor 
\State
\State $L_{\textit{if}} \gets \textit{argsort}(\textit{Influence}, \textit{top}=k)$ \Comment{Top low influences}
\State $L_{\textit{cf}} \gets \textit{argsort}(\textit{Confidence}, \textit{top}=k)$
\State $H_{\textit{cf}} \gets \textit{argsort}(\textit{Confidence}, \textit{bottom}=k)$
\State
\State $S\gets L_{\textit{if}}\setminus (L_{\textit{cf}}\cup H_{\textit{cf}})$ \Comment{Candidate mislabelled data}
\State 
\If{$\textit{mean}(\textit{Influence}, S)\! >\!\textit{mean}(\textit{Influence}, L_{\textit{if}} \cap L_{\textit{cf}}) $} 
    \State $S\gets\emptyset$
\EndIf
\State \Return S
\end{algorithmic}
\AddNote{top1}{bottom1}{left}{}
\vspace{-4mm}
\end{algorithm}
\vspace{-2mm}
Compared to the self-influence method~\cite{KohL17}, our proposed approach does not require computing the Hessian Vector Product~(HVP) for each training data. Instead, it caches the HVP for the entire training set and computes individual data point influence through simple gradient estimation and dot-product (with the cached HVP). While using the entire training data's HVP may hurt our approach's debugging performance, we empirically show that the performance degradation is negligible.
\begin{figure*}[t!]
     \centering
     \begin{subfigure}[]{0.33\linewidth}
         \centering
         \includegraphics[width=\linewidth]{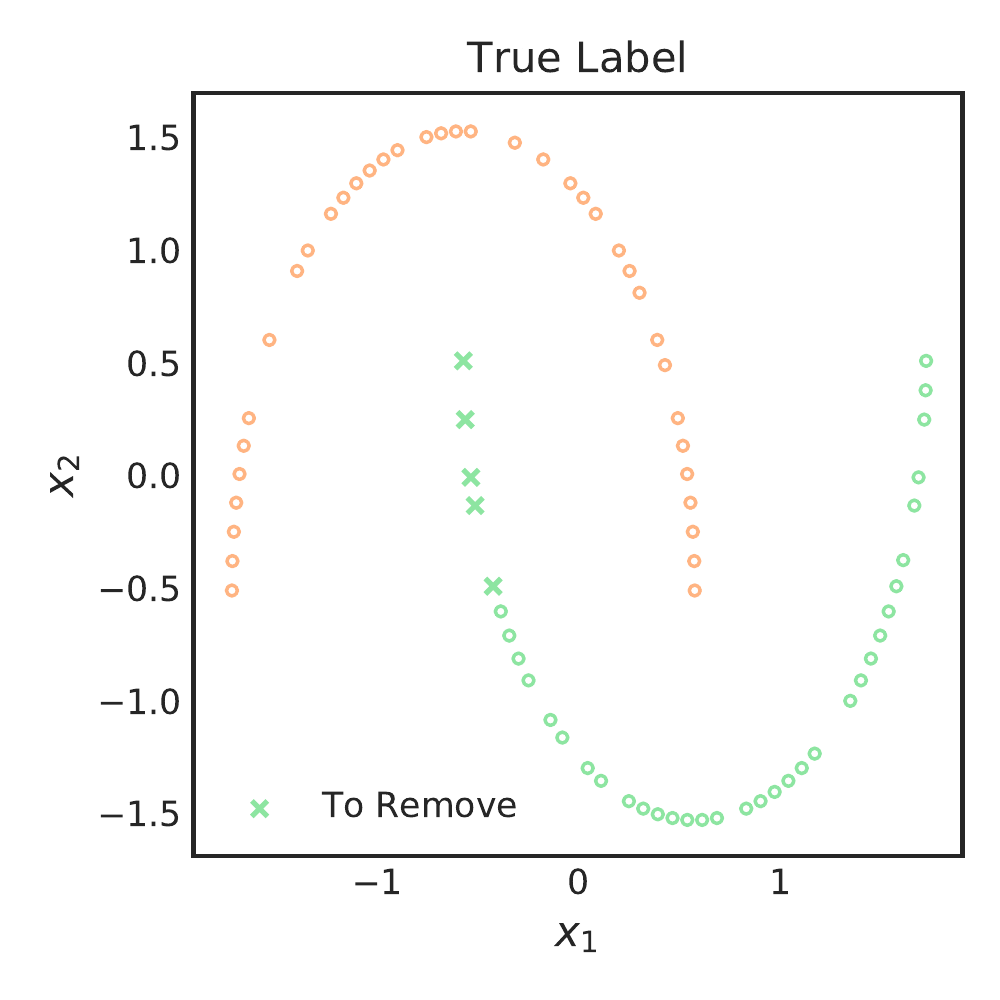}
         \caption{Data}
     \end{subfigure}
     \begin{subfigure}[]{0.33\linewidth}
         \centering
         \includegraphics[width=\linewidth]{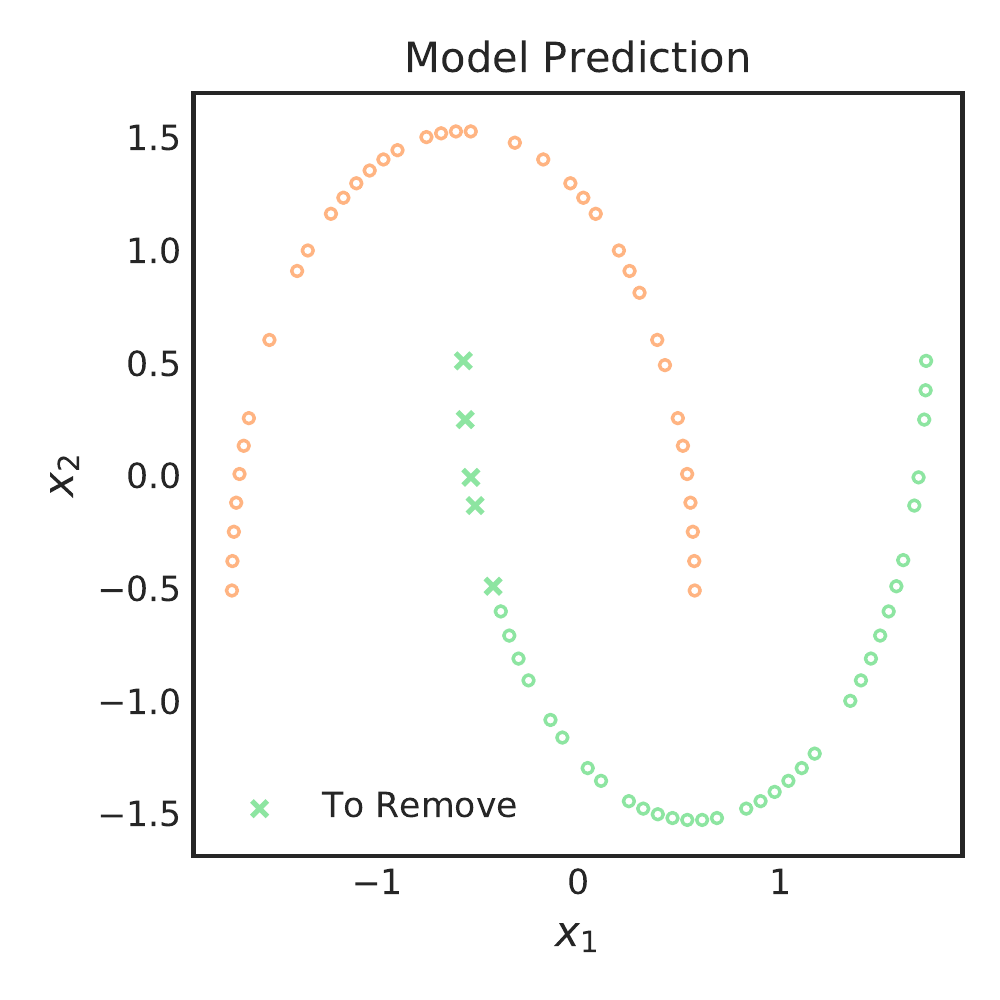}
         \caption{Model prediction before removal}
     \end{subfigure}
     \begin{subfigure}[]{0.33\linewidth}
        \centering
        \includegraphics[width=\linewidth]{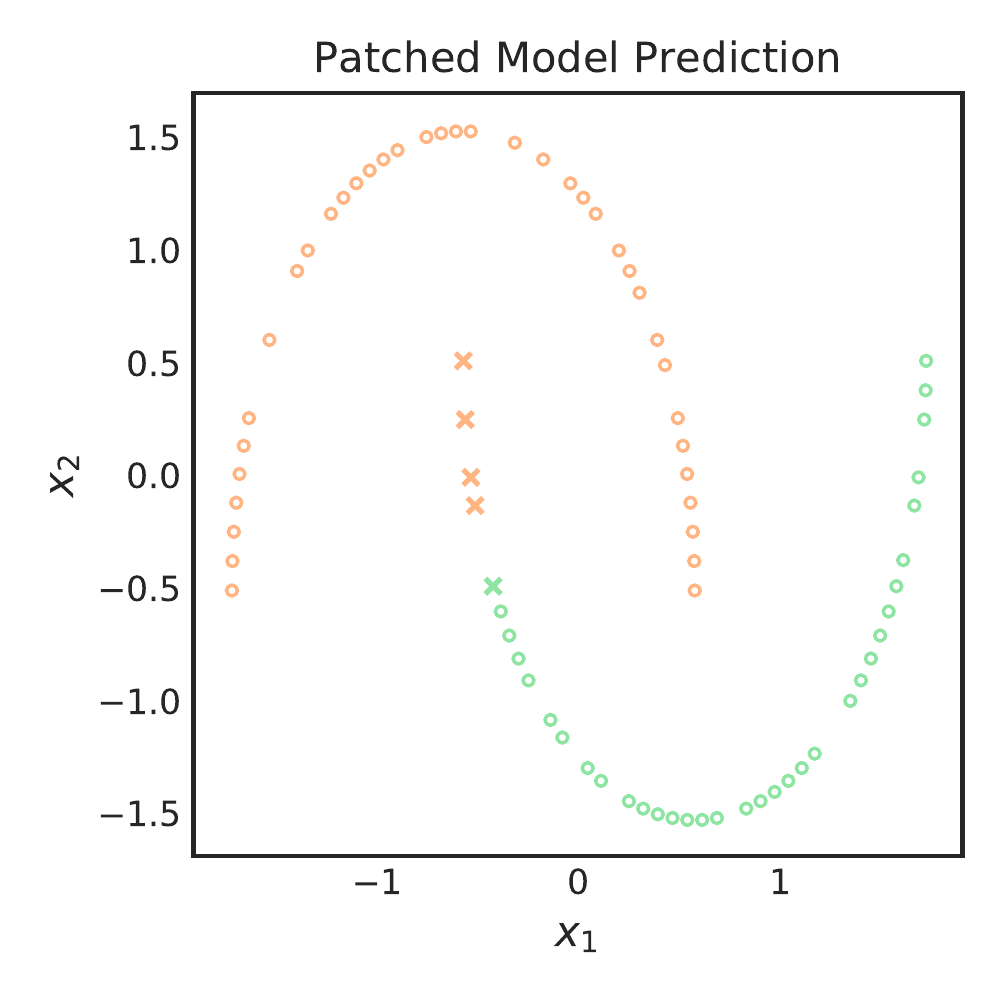}
         \caption{Model prediction after removal}
     \end{subfigure}
    \caption{Removing the training points marked by crosses from the model. As demonstrated in the right plot, PUMA successfully removed the information of all marked points. `x' in the plot shows the data intended to remove. Colors show the class labels.}
    \label{fig:apendix_synthetic_moons}
\end{figure*}

\begin{figure*}[t!]
     \centering
     \begin{subfigure}[]{0.245\linewidth}
         \centering
         \includegraphics[width=0.95\linewidth]{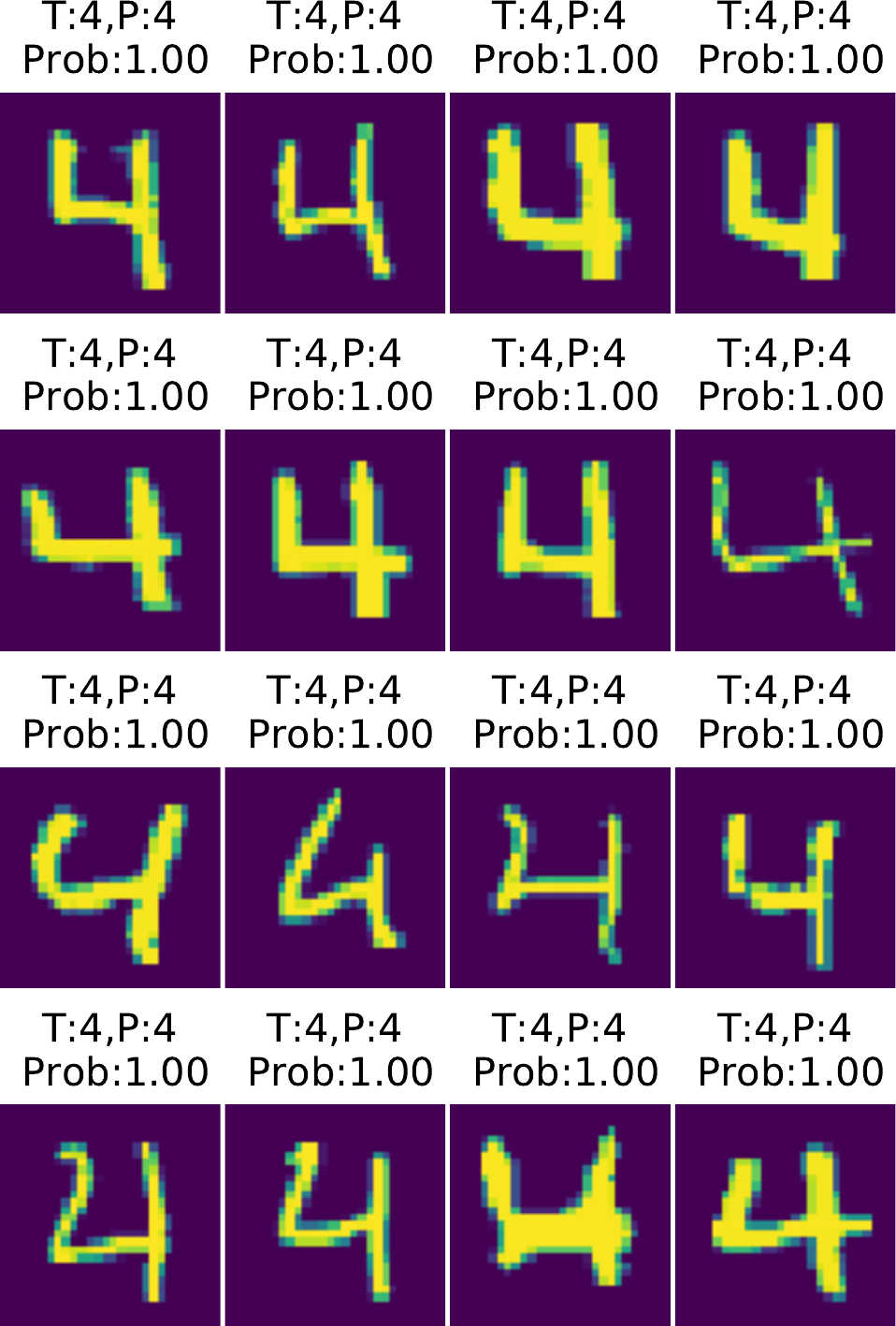}
         \caption{Data marked to remove \\\hspace{\textwidth} (Predictions of original model)}
     \end{subfigure}
     \begin{subfigure}[]{0.245\linewidth}
         \centering
         \includegraphics[width=0.95\linewidth]{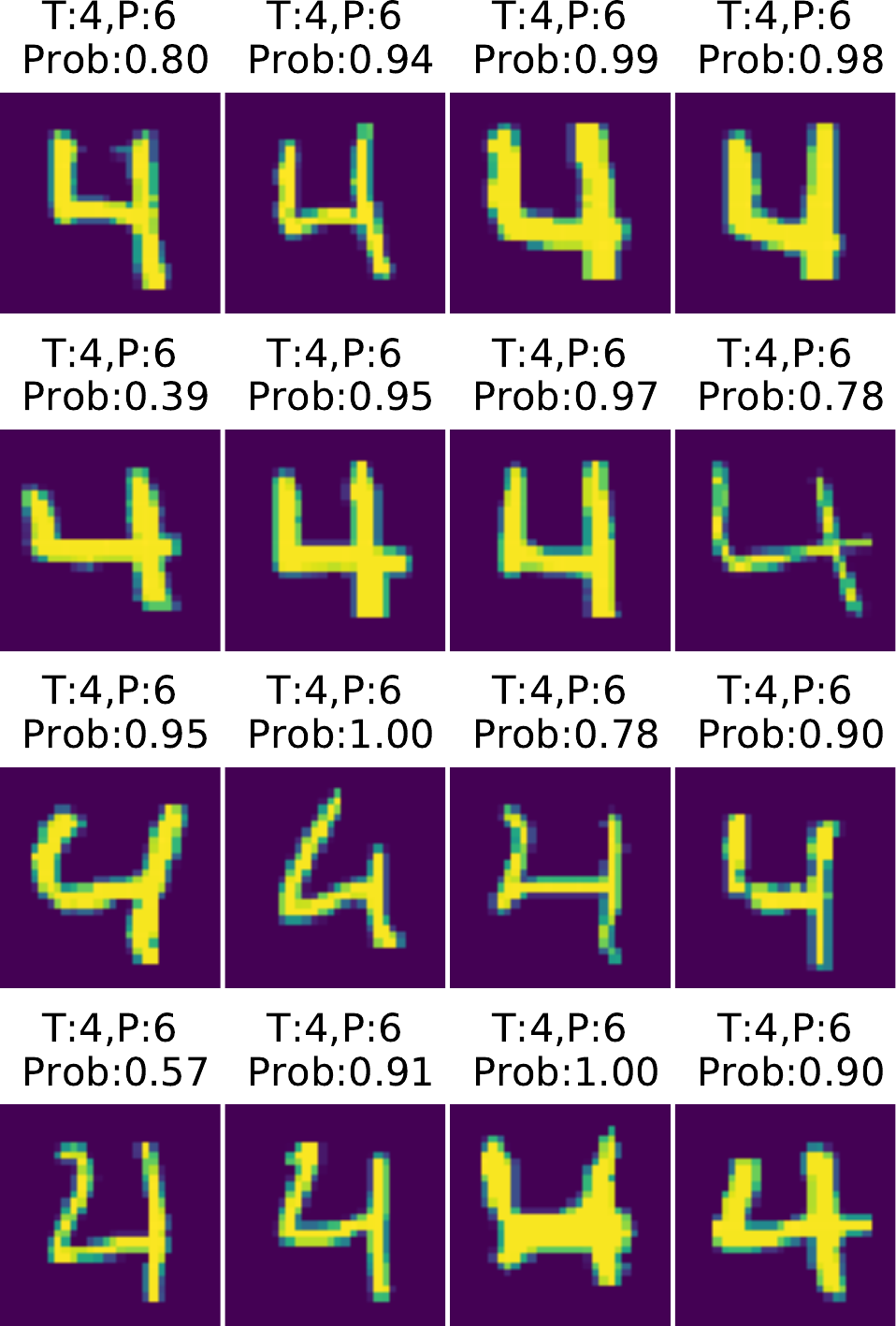}
         \caption{Data marked to remove \\\hspace{\textwidth} (Predictions of augmented model)}
     \end{subfigure}
     \begin{subfigure}[]{0.245\linewidth}
        \centering
        \includegraphics[width=0.95\linewidth]{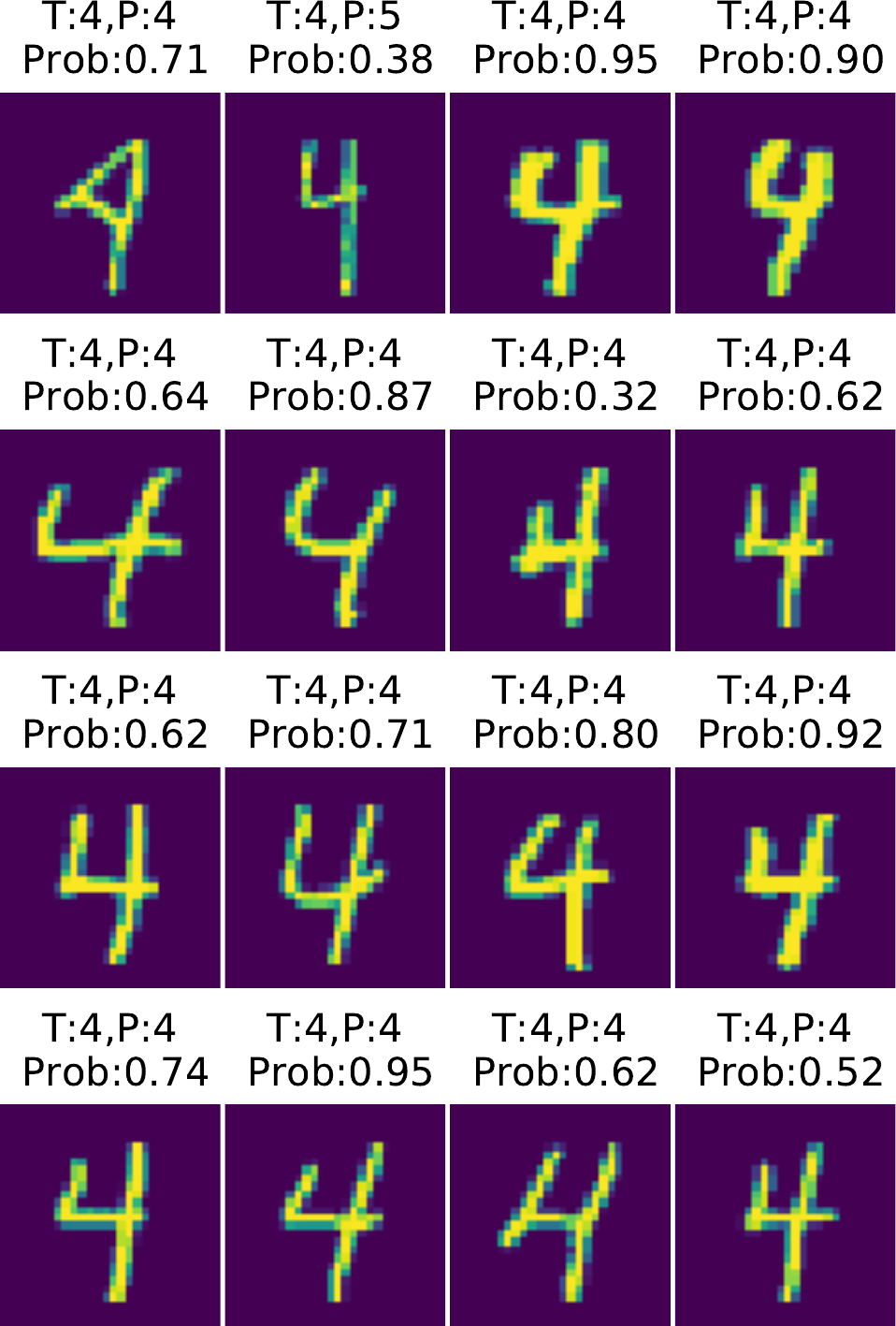}
         \caption{Other data in the same class \\\hspace{\textwidth} (Predictions of augmented model)}
     \end{subfigure}
     \begin{subfigure}[]{0.245\linewidth}
        \centering
        \includegraphics[width=0.95\linewidth]{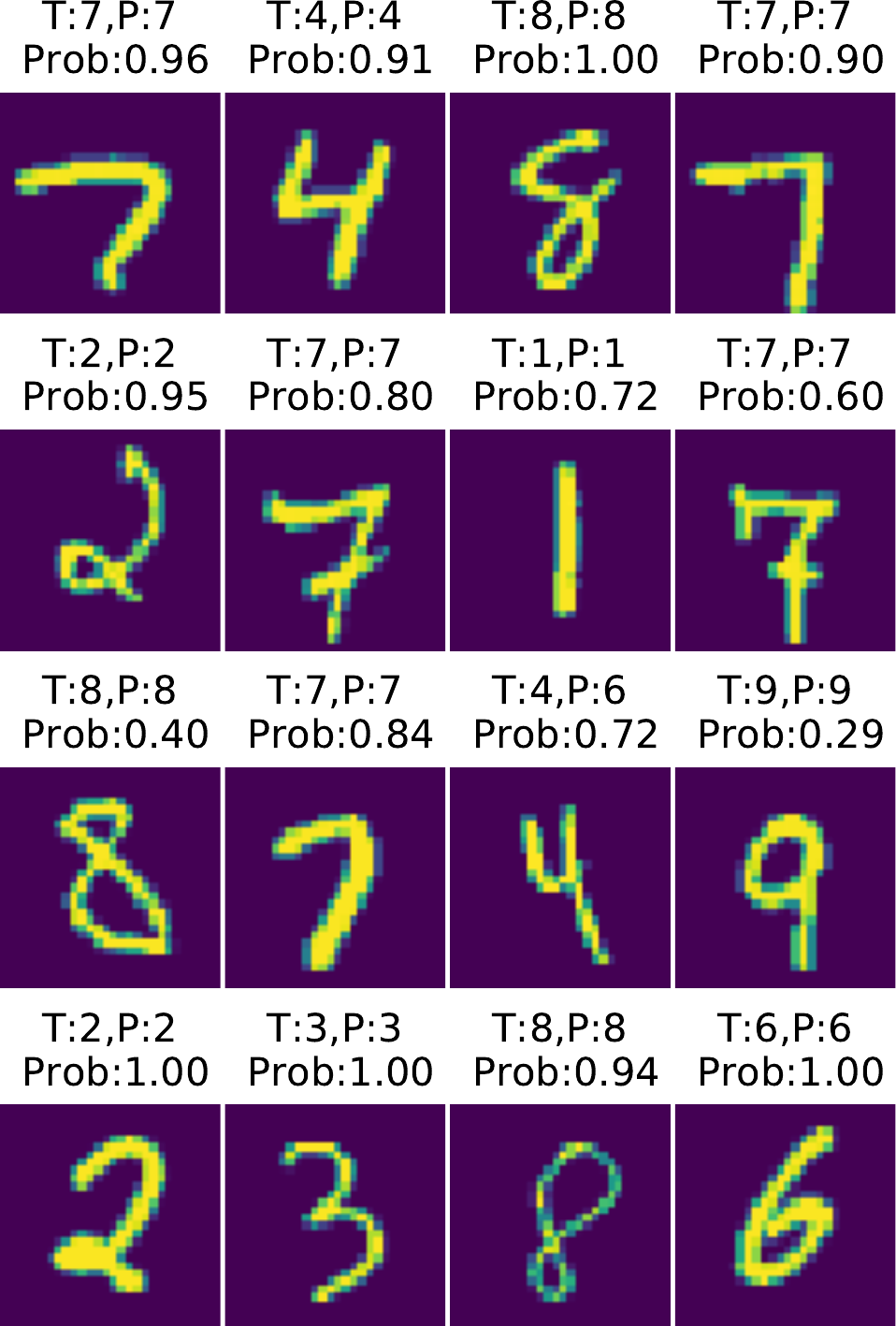}
         \caption{Randomly sampled data \\\hspace{\textwidth} (Predictions of augmented model)}
     \end{subfigure}
    \caption{Removing marked data points that belong to a particular style of digit 4. On top of each digit image, we show the ground truth label (denoted as T), predicted label (denoted as P), and the prediction confidence (denoted as Prob). }
    \label{fig:apendix_mnist}
\end{figure*}
\section{Appendix C: More Demo Cases}
In this section, we show two demonstrative examples which visualize the data removal effect of PUMA.  

Figure~\ref{fig:apendix_synthetic_moons}(a) shows a binary classification task, where data points are distributed as moon shapes. Since there is no noise introduced in the data generation process, the model can make perfect predictions as shown in Figure~\ref{fig:apendix_synthetic_moons}(b). When we remove the data points marked as `x' from the train model (see Figure~\ref{fig:apendix_synthetic_moons}(c)), the top four data points are classified as belonging to the orange class and the bottom one is still classified as the green class. More importantly, the predictions for the other data points remain unchanged. This observation reflects our intuition, where PUMA would leverage remaining data points to stabilize the model's generalization ability by complementing the loss of the removed data points.

Figure~\ref{fig:apendix_mnist} shows a more complex removal case, where we try to remove some data points belonging to digit 4 with a certain writing style. Figure~\ref{fig:apendix_mnist} (a) shows that, before the removal operation, the model predicts the marked data points quite well; all of the predictions are correct with high confidence ($=1.0$).  After the removal operation, we note the marked data points are misclassified with various prediction confidences, as shown in Figure~\ref{fig:apendix_mnist} (b). In contrast, the predictions on other data points of the same class are not seriously affected by the removal operation as their predictions are still correct, as shown in Figure~\ref{fig:apendix_mnist} (c). This demonstrates that the PUMA data removal operation does not false-fully generalize to the other data points in the same class, even if they look similar. Finally, Figure~\ref{fig:apendix_mnist} (d) shows that the removal operation also preserved the prediction accuracy for other inputs.
\begin{table*}[t!]
\caption{Comparison of Model Performance Preservation among Candidate Removal Approaches with Statistics. Value shows accuracy. Higher is better after data removal. Results are collected from 20 times independent experiments. }
\resizebox{\linewidth}{!}{
\begin{tabular}{l|l|clcccc}
\toprule
\multirow{2}{*}{Data Group}&\multirow{2}{*}{Dataset} &\multicolumn{6}{c}{Ordered}\\
\cmidrule(lr){3-8}
&& Original & Approach & 20\% & 40\% & 60\% & 80\% \\
\midrule
\multirow{6}{*}{Synthetic}&\multirow{4}{*}{Radial} & $95.04\pm0.59$ & Retrain Model & $\bm{93.64\pm0.87}$ & $\bm{91.60\pm1.22}$ & $\bm{84.15\pm3.00}$ & $66.24\pm6.42$  \\
& & $80.88\pm3.62$ & SISA & $67.35\pm8.16$ & $61.93\pm7.86$ & $63.57\pm9.53$ & $51.91\pm4.57$ \\
& & $95.04\pm0.59$ & Amnesiac ML & $56.38\pm4.85$ & $54.75\pm4.46$ & $53.53\pm4.68$ & $50.54\pm1.60$ \\
& & $94.97\pm0.64$ & PUMA & $68.97\pm9.47$ & $69.60\pm4.88$ & $67.99\pm5.62$ & $\bm{70.77\pm7.61}$ \\
\cmidrule(lr){2-8}
&\multirow{4}{*}{Rectangular} & $62.00\pm0.81$ & Retrain Model & $\bm{61.20\pm 1.08}$ & $\bm{60.35\pm1.10}$ & $\bm{55.80\pm3.75}$ & $54.25\pm3.86$ \\
& & $55.60\pm5.04$ & SISA & $55.90\pm4.97$ & $48.30\pm12.39$ & $30.10\pm7.78$ & $29.55\pm5.71$ \\
& & $62.00\pm0.81$ & Amnesiac ML & $46.60\pm9.34$ & $43.85\pm8.37$ & $43.45\pm9.73$ & $39.15\pm4.29$ \\
& & $61.85\pm0.74$ & PUMA & $55.25\pm8.28$ & $56.30\pm6.11$ & $53.85\pm9.64$ & $\bm{61.70\pm6.80}$ \\
\midrule
\multirow{6}{*}{\shortstack{Tabular\\ (UCI)}}&\multirow{4}{*}{German} & $71.52\pm0.49$ & Retrain Model & $\bm{70.56\pm0.86}$ & $70.12\pm0.80$ & $70.11\pm0.73$ & $70.00\pm0.00$ \\
& & $70.00\pm0.00$ & SISA & $70.00\pm0.00$ & $70.00\pm0.00$ & $68.96\pm3.28$ & $66.16\pm5.23$ \\
& & $71.52\pm0.49$ & Amnesiac ML & $68.52\pm5.26$ & $64.40\pm10.80$ & $66.24\pm8.60$ & $64.03\pm12.27$ \\
& & $71.47\pm0.50$ & PUMA & $69.08\pm3.19$ & $\bm{70.72\pm0.77}$ & $\bm{70.64\pm0.73}$ & $\bm{70.72\pm1.07}$ \\
\cmidrule(lr){2-8}
&\multirow{4}{*}{Breast Cancer} & $96.45\pm0.45$ & Retrain Model & $\bm{96.62\pm0.42}$ & $\bm{96.11\pm0.59}$ & $96.00\pm0.54$ & $94.85\pm1.92$ \\
& & $91.31\pm1.84$ & SISA & $89.20\pm1.67$ & $88.91\pm5.61$ & $80.68\pm12.00$ & $52.62\pm26.43$ \\
& & $96.45\pm0.45$ & Amnesiac ML & $96.05\pm1.71$ & $95.82\pm1.54$ & $95.25\pm2.85$ & $82.28\pm11.89$ \\
& & $96.39\pm0.47$ & PUMA & $96.17\pm1.42$ & $95.88\pm1.67$ & $\bm{96.22\pm0.61}$ & $\bm{96.62\pm0.62}$ \\
\midrule
\multirow{3}{*}{Image}&\multirow{4}{*}{MNIST} & $97.58\pm0.48$ & Retrain Model & $\bm{97.28\pm0.16}$ & $\bm{96.72\pm0.56}$ & $95.76\pm0.73$ & $93.48\pm0.47$ \\
& & $95.89\pm0.24$ & SISA & $95.80\pm0.36$ & $95.67\pm0.57$ & $94.78\pm0.46$ & $89.86\pm0.39$ \\
& & $97.44\pm0.43$ & Amnesiac ML & $9.44\pm1.12$ & $9.84\pm1.42$ & $9.56\pm1.25$ & $9.36\pm1.24$ \\
& & $97.60\pm0.37$ & PUMA & $96.70\pm0.99$ & $96.66\pm0.48$ & $\bm{97.17\pm0.33}$ & $\bm{97.16\pm0.26}$ \\
\bottomrule
\end{tabular}}
\resizebox{\linewidth}{!}{
\begin{tabular}{l|l|clcccc}
\toprule
\multirow{2}{*}{Data Group}&\multirow{2}{*}{Dataset}&\multicolumn{6}{c}{Random}\\
\cmidrule(lr){3-8}
& & Original & Approach & 20\% & 40\% & 60\% & 80\%\\
\midrule
\multirow{6}{*}{Synthetic}&\multirow{4}{*}{Radial} & $95.89\pm0.88$ & Retrain Model & $\bm{93.97\pm1.01}$ & $\bm{90.94\pm1.25}$ & $\bm{82.58\pm4.87}$ & $66.51\pm6.61$ \\
& & $75.62\pm5.55$ & SISA & $64.71\pm7.29$ & $64.35\pm8.78$ & $54.80\pm3.65$ & $54.77\pm5.70$\\
& & $95.88\pm0.88$ & Amnesiac ML & $49.08\pm2.31$ & $48.95\pm2.20$ & $48.95\pm2.20$ & $48.95\pm2.20$\\
& & $95.82\pm0.86$ & PUMA & $72.44\pm7.05$ & $73.22\pm7.82$ & $71.82\pm7.39$ & $\bm{76.02\pm8.43}$\\
\cmidrule(lr){2-8}
&\multirow{4}{*}{Rectangular} & $65.00\pm0.23$ & Retrain Model & $\bm{64.70\pm1.00}$ & $\bm{64.50\pm1.54}$ & $62.30\pm2.49$ & $58.65\pm3.55$ \\
& & $56.50\pm0.00$ & SISA & $56.50\pm0.00$ & $56.50\pm0.00$ & $56.55\pm0.15$ & $56.90\pm1.44$ \\
& & $65.00\pm0.23$ & Amnesiac ML & $35.40\pm19.08$ & $35.40\pm19.08$ & $35.40\pm19.08$ & $35.40\pm19.08$ \\
& & $64.95\pm0.15$ & PUMA & $59.90\pm4.53$ & $62.05\pm2.74$ & $\bm{62.55\pm2.49}$ & $\bm{64.80\pm1.60}$ \\
\midrule
\multirow{6}{*}{\shortstack{Tabular\\ (UCI)}}&\multirow{4}{*}{German} & $75.16\pm0.66$ & Retrain Model & $\bm{74.88\pm0.95}$ & $\bm{73.24\pm0.66}$ & $72.47\pm0.88$ & $70.00\pm0.18$ \\
& & $70.00\pm0.00$ & SISA & $70.00\pm0.00$ & $70.00\pm0.00$ & $70.00\pm0.00$ & $70.00\pm0.00$ \\
& & $75.16\pm0.66$ & Amnesiac ML & $36.24\pm7.50$ & $36.28\pm7.46$ & $35.72\pm7.69$ & $35.72\pm7.69$ \\
& & $75.12\pm0.61$ & PUMA & $70.96\pm3.62$ & $\bm{73.24\pm1.39}$ & $\bm{74.44\pm0.66}$ & $\bm{74.28\pm1.29}$\\
\cmidrule(lr){2-8}
&\multirow{4}{*}{Breast Cancer} & $96.00\pm0.00$ & Retrain Model & $\bm{95.82\pm0.27}$ & $\bm{95.54\pm0.36}$ & $\bm{95.65\pm0.39}$ & $95.20\pm0.67$  \\
& & $92.28\pm2.24$ & SISA & $91.60\pm2.95$ & $88.05\pm4.57$ & $88.22\pm4.20$ & $87.88\pm5.23$ \\
& & $96.00\pm0.00$ & Amnesiac ML & $35.20\pm20.33$ & $30.51\pm18.97$ & $30.51\pm18.97$ & $30.51\pm18.97$ \\
& & $96.00\pm0.00$ & PUMA & $95.08\pm1.94$ & $94.91\pm2.29$ & $95.25\pm1.76$ & $\bm{95.54\pm0.24}$ \\
\midrule
\multirow{3}{*}{Image}&\multirow{4}{*}{MNIST} & $97.99\pm0.36$ & Retrain Model & $\bm{97.72\pm0.29}$ & $97.16\pm0.42$ & $96.60\pm0.74$ & $93.98\pm0.72$\\
& & $95.66\pm0.41$ & SISA & $93.47\pm0.68$ & $90.63\pm1.51$ & $78.06\pm3.28$ & $59.83\pm10.02$\\
& & $98.06\pm0.37$ & Amnesiac ML & $10.39\pm1.32$ & $10.39\pm1.32$ & $10.39\pm1.32$ & $10.39\pm1.32$\\
& & $97.97\pm0.34$ & PUMA & $97.42\pm0.66$ & $\bm{97.58\pm0.36}$ & $\bm{97.60\pm0.58}$ & $\bm{97.61\pm0.44}$\\
\bottomrule
\end{tabular}}
\label{table:performance_perservaction_with_statistics}
\end{table*}

\section{Appendix D: Results with Statistics}
Here we show the full table of our experiment results with statistics that were omitted in the main paper.

\subsection{Performance Preservation with Statistics}

Table~\ref{table:performance_perservaction_with_statistics} shows the performance preservation table. The mean values are exactly the same as presented in the main paper. Statistics show standard deviation of 10 runs. Here, we note that Amnesiac ML often shows a large variance compared to the other approaches, even when the marked data points are intentionally organized into a small number of batches (see results in Ordered scenario). In addition, we also observe that, on the small datasets, training multiple sub-models (as SISA does) often results in bad performance compared with single model approaches (Retrain, Amnesiac ML, and PUMA).

\begin{table*}[t]
\caption{Comparison of Membership Attack after Data Removal Operation. Value shows percentage of removed data that is identified as training data. Lower values in the table show better performance of removal.}
\resizebox{\linewidth}{!}{
\begin{tabular}{l|l|cc|cc|cc|cc}
\toprule
\multirow{4}{*}{\shortstack{Data\\ Group}} & \multirow{4}{*}{Dataset} & \multicolumn{8}{c}{Ordered} \\
\cmidrule(lr){3-10}
&&\multicolumn{2}{c|}{Retrain Model}&\multicolumn{2}{c|}{SISA}&\multicolumn{2}{c|}{Amnesiac ML}&\multicolumn{2}{c}{PUMA}\\
\cmidrule(lr){3-10}
&&Before&After&Before&After&Before&After&Before&After\\
\midrule
\multirow{2}{*}{Synthetic}&Radial&$100.00\pm0.00$&$100.00\pm0.00$&$100.00\pm0.00$&$100.00\pm0.00$&$100.00\pm0.00$&$\bm{0.00\pm0.00}$&$100.00\pm0.00$&$5.31\pm10.38$\\
&Rectangular&$100.00\pm0.00$&$91.65\pm6.20$&$83.18\pm19.76$&$83.18\pm19.76$&$100.00\pm0.00$&$\bm{33.33\pm0.00}$&$100.00\pm0.00$&$36.66\pm29.18$\\
\midrule
\multirow{2}{*}{Tabular}&German&$100.00\pm0.00$&$77.12\pm1.38$&$100.00\pm0.00$&$100.00\pm0.00$&$100.00\pm0.00$&$0.00\pm0.00$&$100.00\pm0.00$&$\bm{3.42\pm6.90}$\\
&Breast Cancer&$100.00\pm0.00$&$100.00\pm0.00$&$87.50\pm8.83$&$87.50\pm8.83$&$100.00\pm0.00$&$100.00\pm0.00$&$100.00\pm0.00$&$\bm{56.25\pm13.50}$\\
\midrule
Image&MNIST&$100.00\pm0.00$&$100.00\pm0.00$&$100.00\pm0.00$&$100.00\pm0.00$&$100.00\pm0.00$&$100.00\pm0.00$&$100.00\pm0.00$&$\bm{0.00\pm0.00}$\\
\bottomrule
\end{tabular}}
\resizebox{\linewidth}{!}{
\begin{tabular}{l|l|cc|cc|cc|cc}
\toprule
\multirow{4}{*}{\shortstack{Data\\ Group}} & \multirow{4}{*}{Dataset} & \multicolumn{8}{c}{Random} \\
\cmidrule(lr){3-10}
&&\multicolumn{2}{c|}{Retrain Model}&\multicolumn{2}{c|}{SISA}&\multicolumn{2}{c|}{Amnesiac ML}&\multicolumn{2}{c}{PUMA}\\
\cmidrule(lr){3-10}
&&Before&After&Before&After&Before&After&Before&After\\
\midrule
\multirow{2}{*}{Synthetic}&Radial&$100.00\pm0.00$&$52.36\pm3.86$&$100.00\pm0.00$&$37.00\pm17.88$&$100.00\pm0.00$&$50.00\pm0.00$&$100.00\pm0.00$&$\bm{1.18\pm1.96}$\\
&Rectangular&$100.00\pm0.00$&$67.07\pm5.29$&$98.50\pm3.37$&$94.00\pm12.86$&$100.00\pm0.00$&$86.20\pm0.00$&$100.00\pm0.00$&$\bm{20.00\pm23.30}$\\
\midrule
\multirow{2}{*}{Tabular}&German&$94.44\pm0.00$&$84.44\pm2.34$&$100.00\pm0.00$&$98.81\pm8.38$&$94.44\pm0.00$&$93.33\pm3.51$&$85.18\pm2.22$&$\bm{2.22\pm7.13}$\\
&Breast Cancer&$100.00\pm0.00$&$100.00\pm0.00$&$90.00\pm7.90$&$73.75\pm27.29$&$100.00\pm0.00$&$87.50\pm0.00$&$100.00\pm0.00$&$\bm{71.25\pm25.71}$\\
\midrule
Image&MNIST&$100.00\pm0.00$&$100.00\pm0.00$&$100.00\pm0.00$&$100.00\pm0.00$&$100.00\pm0.00$&$100.00\pm0.00$&$100.00\pm0.00$&$\bm{72.00\pm13.03}$\\
\bottomrule
\end{tabular}}
\label{table:membership_attack_with_statistics}
\end{table*}

\begin{table*}[t]
\centering
\caption{Comparison of Running Time (in Seconds). Lower values in the table show better performance to the mislabelled data debugging. We omit the statistic in this table for saving space. Please refer to appendix for statistics.}
\resizebox{0.7\linewidth}{!}{
\begin{tabular}{lccccc}
\toprule
\multirow{2}{*}{Data} & \multicolumn{5}{c}{Approach}  \\
\cmidrule(lr){2-6}
 & GShapley & NTK & SelfInfluence & RSP & PUMA \\
\midrule
Two Moons & $90.37 \pm 2.50$ & $22.65 \pm 0.62$ & $1562.93 \pm 113.78$ & $17.14 \pm 0.17$ & $\bm{7.61 \pm 0.08}$ \\
Spiral & $78.47 \pm 0.62$ & $19.91 \pm 0.23$ & $1464.01 \pm 5.32$ & $16.51 \pm 0.13$ & $\bm{7.44 \pm 0.11}$ \\
Radial & $82.99 \pm 0.29$ & $21.60 \pm 0.42$ & $1563.53 \pm 2.54$ & $17.76 \pm 0.27$ & $\bm{7.76 \pm 0.07}$ \\
Rectangulars & $78.04 \pm 0.79$ & $20.23 \pm 0.38$ & $1480.12 \pm 9.33$ & $16.81 \pm 0.26$ & $\bm{7.29 \pm 0.08}$ \\
% German Credit & $54.30 \pm 6.13$ & $7.95 \pm 0.15$ & $687.02 \pm 14.31$ & $7.67 \pm 3.04$ & $7.35 \pm 0.27$ \\
% Breast Cancer & $14.53 \pm 2.72$ & $3.28 \pm 0.05$ & $218.92 \pm 10.73$ & $2.42 \pm 0.07$ & $1.25 \pm 0.02$ \\

\bottomrule
\end{tabular}}
\label{table:run_time_debugging_with_statistics}
\end{table*}

\subsection{Removal Performance with Statistics}
Table~\ref{table:membership_attack_with_statistics} shows the data removal performance table with statistics. As the data points marked to be removed are randomly selected in each experiment, we observe large variances in the table. However, since the mean values reported in the table are dramatically different, we don't observe an overlap among the statistics of the results.

\subsection{Execution Time for Debugging}
Table~\ref{table:run_time_debugging_with_statistics} shows the execution time for mislabelled data debugging with statistics. 
\section{Appendix E: Model Calibration Application}
As described in the main paper, PUMA preserves model performance while removing and reweighting the contributions of the training data points. The performance can be measured by any differentiable performance criterion $\mathcal{C}$. In this example, we define the performance criterion $\mathcal{C}$ as an Expected Calibration Error~(ECE)~\cite{guo2017calibration}.

In the following context, we first use PUMA's data debugging ability to identify the problematic data points causing mis-calibration. We then use PUMA's data removal functionality to update the model parameters to get a better calibrated model.
\subsection{Identifying Problematic Data Points}
To identify the problematic data points causing mis-calibration, we need to modify the PUMA debugging algorithm to filter the problematic data points into three categories, namely over-confidence, over-uncertain, and other-noise. Specifically, we make the following adjustment on the Algorithm~\ref{alg:puma_debugging} to meet the requirement.
\vspace{-2mm}
\begin{algorithm}[h]
\caption{PUMA Problematic Data Debugging}\label{alg:puma_debugging_adjusted}
\begin{algorithmic}
\Require Original model $\theta_{\textit{org}}$, Training objective $\mathcal{J}_{\textit{org}}$, Performance criterion $\mathcal{C}$, Whole training set $D_{\textit{tn}}$, Number of problematic data to return $k$
\Ensure $\mathcal{C}$ is differentiable
\State $\cdots$
\State $L_{\textit{if}} \gets \textit{argsort}(\textit{Influence}, \textit{top}=k)$ \Comment{Top low influences}
\State $L_{\textit{cf}} \gets \textit{argsort}(\textit{Confidence}, \textit{top}=k)$
\State $H_{\textit{cf}} \gets \textit{argsort}(\textit{Confidence}, \textit{bottom}=k)$
\State
\State $S_1\gets L_{\textit{if}}\cap H_{\textit{cf}}$ \Comment{Over-confident}
\State $S_2\gets L_{\textit{if}}\cap L_{\textit{cf}}$ \Comment{Over-uncertain}
\State $S_3\gets L_{\textit{if}}\setminus (L_{\textit{cf}}\cup H_{\textit{cf}})$ \Comment{Other-noise}
\State 

\State \Return $S_1$, $S_2$, $S_3$
\end{algorithmic}
\end{algorithm}
\vspace{-2mm}

When the performance criterion is ECE, running the above algorithm, we can identify the three categories of data which negatively influence the predictive uncertainty estimation of the model~(increasing ECE). 

\begin{figure}[t!]
     \centering
     \begin{subfigure}[]{0.495\linewidth}
         \centering
         \includegraphics[width=\linewidth]{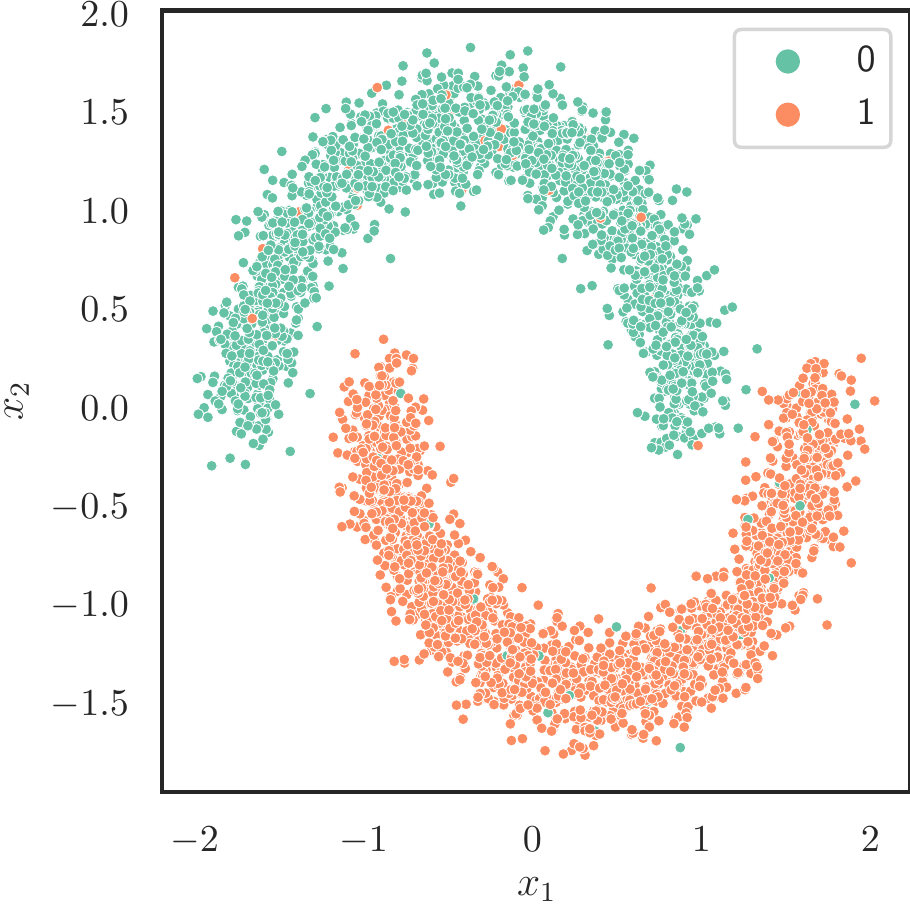}
         \caption{Noisy Data}
     \end{subfigure}
     \begin{subfigure}[]{0.495\linewidth}
         \centering
         \includegraphics[width=\linewidth]{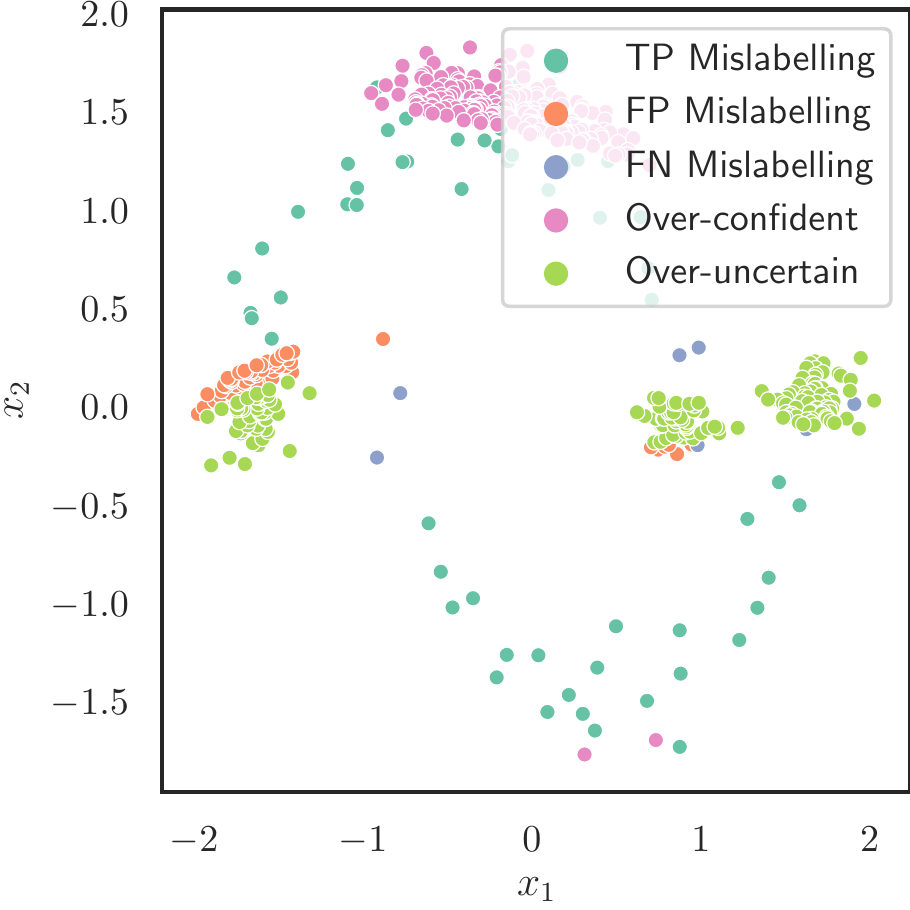}
         \caption{Problem Highlight}
     \end{subfigure}
    \caption{Problematic data points discovered by PUMA. The problems are categorized into three categories. For the prediction of mislabelled data points, we show the prediction's True Positive (TP), False Positive (FP), and False Negative (FN).}
    \label{fig:two_moon_debugging}
\vspace{-3mm}
\end{figure}

Figure~\ref{fig:two_moon_debugging} shows our experiment results on the synthetic data. The data is heavily corrupted two-moon synthetic data with 5000 data points. Among the data points, 100 data points' labels are randomly flipped. We run PUMA with ECE loss as the performance metric $\mathcal{C}$ to debug the trained model on the corrupted data. Results show that PUMA can fairly accurately identify the various problems in the model (i.e. the three data categories). Note, unlike the mislabelled data points, the identified over-confident and over-uncertain predictions are part of the model problems, not a training data quality issue. Hence, in Figure~\ref{fig:two_moon_debugging}(b), the over-confident predictions are not symmetric in the two classes for this particular model.

\begin{figure}[t!]
     \centering
     \begin{subfigure}[]{0.495\linewidth}
         \centering
         \includegraphics[width=\linewidth]{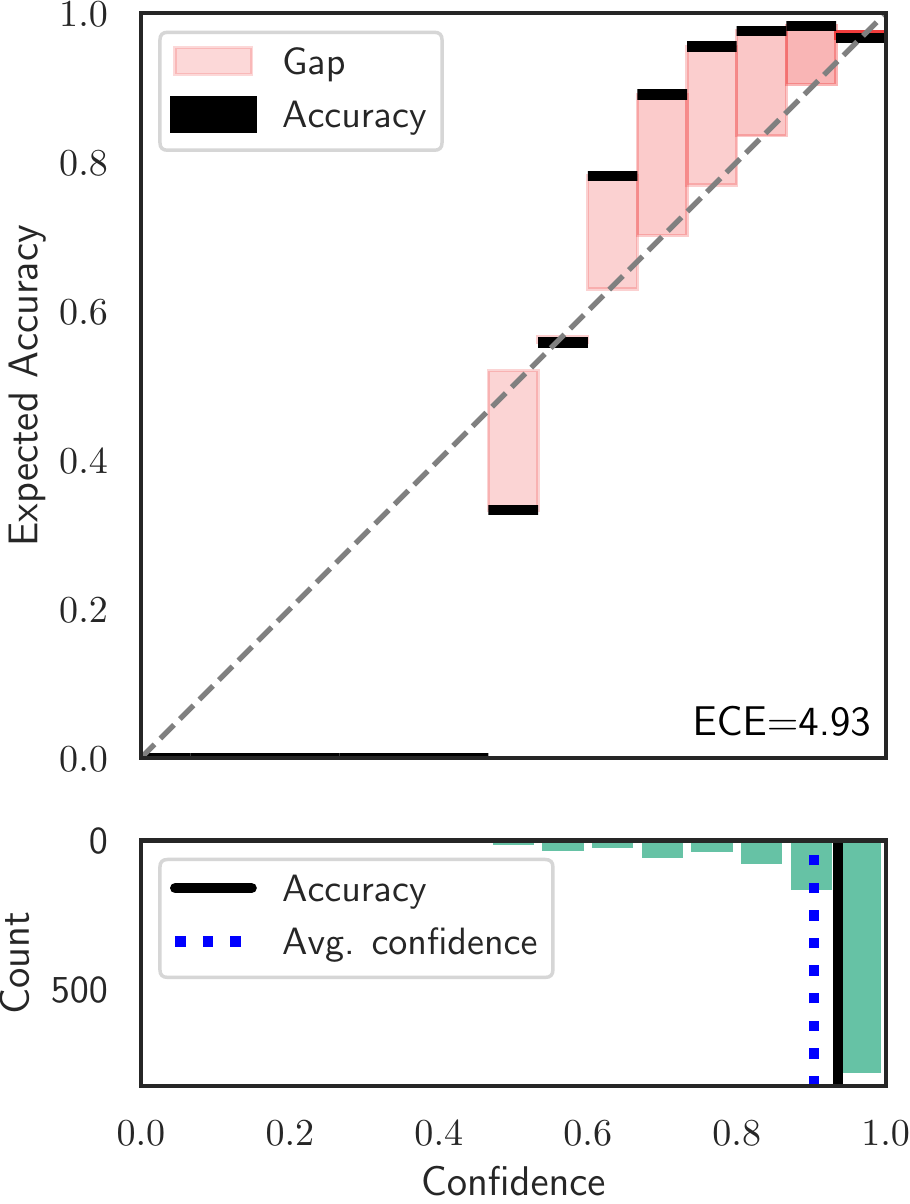}
         \caption{ECE Before Patching}
     \end{subfigure}
     \begin{subfigure}[]{0.495\linewidth}
         \centering
         \includegraphics[width=\linewidth]{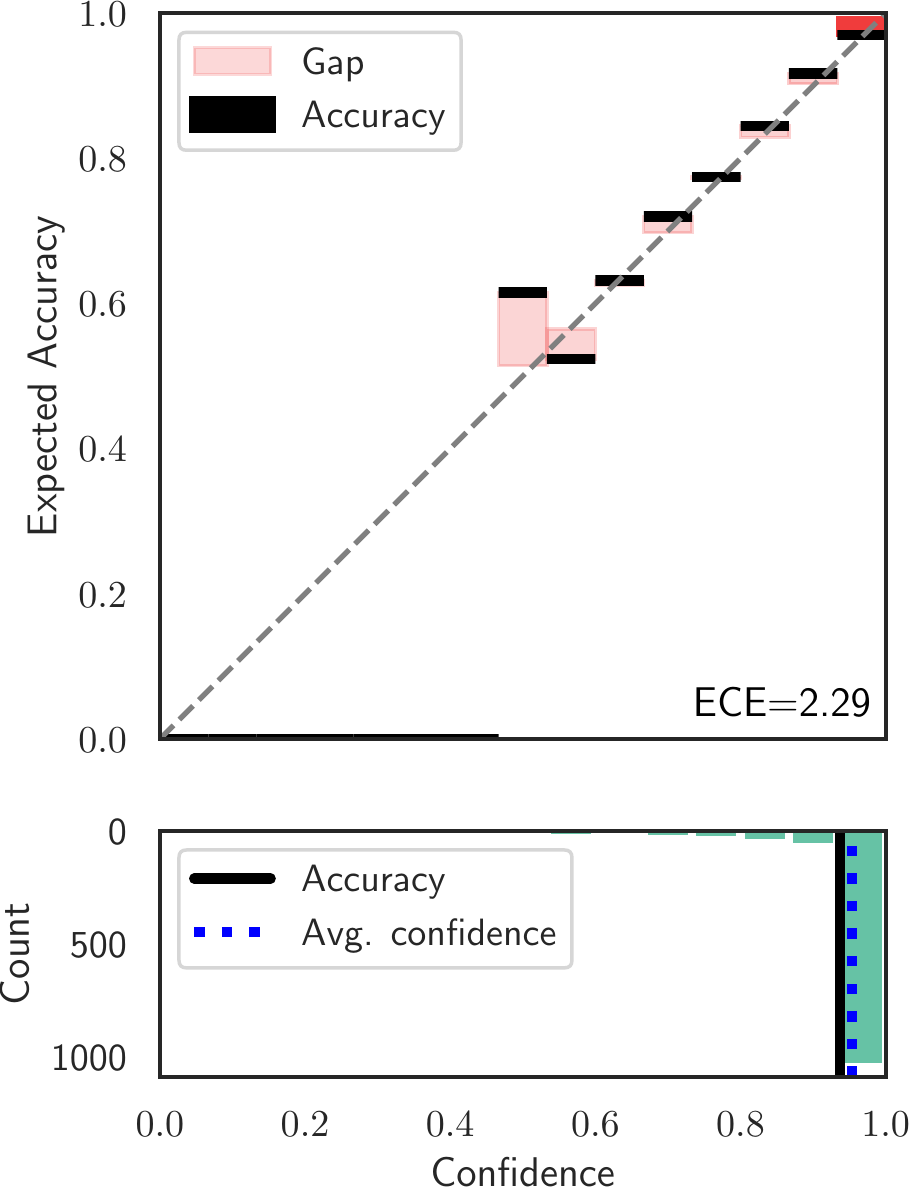}
         \caption{ECE After Patching}
     \end{subfigure}
    \caption{Expected Calibration Error~(ECE) before and after PUMA model patching/augmentation. For most of prediction confidence bins, we observe the better calibration after PUMA model augmentation.}
    \label{fig:two_moon_patching}
    \vspace{-5mm}
\end{figure}

\subsection{Model Patching with PUMA}
Now, if we set $D_{\textit{mk}}\stackrel{\text{def}}{=}S_1\cup S_3$ and $D_{\textit{up}}\stackrel{\text{def}}{=}S_2$ and run the PUMA data removal algorithm (see Algorithm~\ref{alg:puma_removal}) with small learning rate $\eta$ (E.g. $\eta\leq 1e-4$), we expect to see that the ECE loss of the model is reduced after updating the model with PUMA.

Figure~\ref{fig:two_moon_patching} shows the model augmentation results. The results reflect our expectation, since the ECE loss of the updated model is dramatically reduced.

% nothing in here

\section{Appendix F: Discussion}
\subsection{Purging Data vs. Data Characteristic Removal}
Complete removal of data points from a trained model is barely possible since the model training procedure is usually complex and it mixes information from all training data points with mutual dependence. In particular, training process of the modern deep learning often uses momentum-enabled optimization algorithms to avoid getting stuck in a local minimum. This makes the decomposition of the contribution of each training data point in learned model parameters hard. In addition, approaches such as batch normalization and weight decay could make the data contribution estimation even more intractable. Hence, in this paper, we do not intend to completely purge the effects of the training data points. Indeed, the only possible solution for removing data from a model is probably retraining the model from scratch with the remaining data points.

The main goal of this work is to remove the identifiable characteristics of the marked data points such that 1) their negative impact on the model can be mitigated and 2) the marked data points cannot be retrieved or identified by the adversaries. The fundamental difference between our goal and removing/purging data is that we want to keep the general properties of all data points (including the marked ones) that are positively influential to the model such that the model's performance is preserved. This technique is particularly useful when the models under service are required to respond quickly to data removal requests, so that a full model retraining can be conducted offline within certain period of time. Indeed, the immediate response to data removal requests in online service is increasingly important.

\subsection{Privacy Protection vs Data Characteristic Removal}
Privacy protection is a very important research field with many well defined criteria. However, PUMA is not designed to satisfy those criteria (e.g. information leakage~\cite{hitaj2017deep,ateniese2015hacking}, re-identification attack~\cite{yang2021learning}, reconstruction attack~\cite{lyu2021novel}, tracing attack~\cite{homer2008resolving}, model inversion~\cite{fredrikson2015model}). Indeed, this paper does not aim to solve privacy preservation problems. While we use the performance of the membership attack as a metric in the paper, it is meant to show the effectiveness of removing data characteristics only. 

Actually, we note that PUMA could be used in various application scenarios such as improving the prediction calibration of a model. As shown in previous examples in Appendix E, by removing data which causes over-confidence and upweighting over-uncertain data points optimally, PUMA can adjust a model's prediction uncertainty to fit calibration requirements in certain applications. This application does not involve privacy protection but is very useful for cascade models or multi-stage models, where outputs of upstream models are expected to be well calibrated.

\end{document}